\documentclass{article}     

\usepackage{arxiv}
\usepackage[utf8]{inputenc} 
\usepackage[T1]{fontenc}    
\usepackage{hyperref}       
\usepackage{url}            
\usepackage{booktabs}       
\usepackage{amsfonts}       
\usepackage{nicefrac}       
\usepackage{microtype}      
\usepackage{lipsum}
\usepackage{graphicx}
\usepackage{caption}
\usepackage{subcaption}
\usepackage{multirow}
\usepackage{amsmath,amsfonts}
\usepackage{tensor}
\usepackage{multirow,multicol}
\usepackage{orcidlink}
\usepackage[capitalise]{cleveref}
\usepackage{wrapfig}
\usepackage{siunitx}
\title{One-DoF Robotic Design of Overconstrained Limbs with Energy-Efficient, Self-Collision-Free Motion}
\author{
    Yuping Gu \\
    Southern University of Science and Technology\\
    \& The University of Hong Kong\\
    \And
    Bangchao Huang \\
    Southern University of Science and Technology\\
    \AND
    Haoran Sun \\
    Southern University of Science and Technology\\
    \& The University of Hong Kong\\
    \And
    Ronghan Xu, Jiayi Yin, Wei Zhang, Fang Wan \\
    Southern University of Science and Technology\\
    \AND
    Jia Pan* \\
    The University of Hong Kong\\
    \texttt{jpan@cs.hku.hk} \\
    \And
    Chaoyang Song* \\
    asRobotics\\
    \texttt{songcy@ieee.org} \\
}
\begin{document}
\maketitle
\begin{abstract}

    While it is expected to build robotic limbs with multiple degrees of freedom (DoF) inspired by nature, a single DoF design remains fundamental, providing benefits that include, but are not limited to, simplicity, robustness, cost-effectiveness, and efficiency. Mechanisms, especially those with multiple links and revolute joints connected in closed loops, play an enabling factor in introducing motion diversity for 1-DoF systems, which are usually constrained by self-collision during a full-cycle range of motion. This study presents a novel computational approach to designing one-degree-of-freedom (1-DoF) overconstrained robotic limbs for a desired spatial trajectory, while achieving energy-efficient, self-collision-free motion in full-cycle rotations. Firstly, we present the geometric optimization problem of linkage-based robotic limbs in a generalized formulation for self-collision-free design. Next, we formulate the spatial trajectory generation problem with the overconstrained linkages by optimizing the similarity and dynamic-related metrics. We further optimize the geometric shape of the overconstrained linkage to ensure smooth and collision-free motion driven by a single actuator. We validated our proposed method through various experiments, including personalized automata and bio-inspired hexapod robots. The resulting hexapod robot, featuring overconstrained robotic limbs, demonstrated outstanding energy efficiency during forward walking.
    
\end{abstract}
\keywords{
    Computational Design \and Mechanism Synthesis \and Overconstrained Robotics \and Collision Avoidance \and Legged Locomotion
}   
\section{Introduction}
\label{sec:Intro}

    Designing robotic limbs for a specific application is a challenging optimization problem that involves mechanical configuration, workspace, energy efficiency, payload capacity, and numerous other factors~\cite{biswal2021development}. The open-chain limbs are a common choice by serially connecting rotation motors with specific configurations for agile motion and avoiding self-collision, such as the robotic manipulators~\cite{kebria2016kinematic, ollero2021past, riboli2023collision}, quadruped robots~\cite{hutter2016anymal, hoeller2024anymal}, and humanoid robots~\cite{kuindersma2016optimization, koptev2021real, darvish2023teleoperation}. However, the resulting drawbacks include the redundant inertia of moving components and relatively lower payload capacity. In contrast, the closed-chain limbs usually arrange the motors near the body and leverage the extra links and joints in closure form to achieve lower inertia for moving parts and improved end-effector stiffness, which becomes a preferred choice of legged robots~\cite{Kenneally2016Design, park2017high, Kau2019Stanford, liu2020articulated, wu2021design, nasonov2023computational}. The increased number of links and joints makes the design of closed-chain robotic limbs more complex when considering the parametric choices for better performance and collision avoidance for larger workspaces concurrently. The scissor-like joint is widely used in robotic limb design~\cite{he2020mechanism} due to its simplicity and reliability, which ensures an extensive motion range for the joints. At the same time, the collision between links would still significantly reduce the workspace, leaving a research question regarding the simultaneous optimization of the parameters and geometries of closed-chain robotic limbs for specified tasks. The computational design method provides an alternative solution to task-specified robotic limb optimization~\cite{ha2016task, chadwick2020vitruvio}.  Ha et al.~\cite{ha2018computational} present a computational approach to designing the robotic device by combining modular components for high-level motion specifications. On the other hand, it remains challenging to formulate an optimization problem for generating desired trajectories with collision-free closed-chain configurations and reduced actuation. 

    Although robotic arms can achieve these spatial trajectories, they require integrating multiple actuators using complicated control software~\cite{zhou2022review}. Instead, the one-DoF robotic design can perform receptive motion approximately driven by a single actuator. This single-actuator-driven characteristic has several advantages, including ease of control, robustness, low cost, and lightweight design, making it widely used in various engineering applications, such as modern machines~\cite{roussel2018exploratory}, automata~\cite{ceylan2013designing, coros2013computational}, and robotics~\cite{mannhart2020cami}. The RHex robot series provides an instructive insight into robotic limb design by rotating each limb in the sagittal plane and mimicking the behavior of cockroaches~\cite{saranli2001rhex, altendorfer2001rhex, saranli2000design}. The simplicity of the limb design has resulted in a robust robot platform for learning legged locomotion~\cite{prahacs2004towards}. With the development of this series of robots, the limb design converged on a planar four-bar structure with full-cycle motion for dynamic locomotion~\cite{lin2004toward}.

    On the other hand, the planar four-bar mechanism is also widely adopted in other robotic designs~\cite{he2020mechanism}. In contrast, the generalized four-bar linkage (Bennett linkage) still has limited engineering applications as a robotic limb. The early study by Carvalho and Silvestre~\cite{carvalho2016motion} utilized the Bennett linkage for designing a hexapod robot's limb, with each limb comprising an active revolute joint and two additional actuators for overcoming obstacles. However, the proposed design is redundant and only demonstrates the limited advantages of the Bennett limb in a simulation environment. Gu et al.~\cite{Gu2022OverconstrainedCoaxial} investigate the design of the Bennett linkage as a robotic limb with coaxial actuation, enabling omni-directional locomotion. This demonstrates the potential advantage of an overconstrained linkage, which results in a systematic reduction in actuation.

    Furthermore, the Bennett robotic limb has also been validated to be energy-efficient for specific tasks by comparing it with other configurations~\cite{gu2023computational, Sun2024OCLocomotion}. A research gap remains in leveraging overconstrained linkages for energy-efficient robotic limb design with reduced actuation due to their relatively complicated geometric condition and kinematic constraints~\cite{song2012family}. This study aims to develop a computational design model that optimizes the parameters and geometries of overconstrained robotic limbs for efficiently realizing a target trajectory with a single actuator.

\subsection{On Linkage Synthesis}
\label{sec:Intro-LinkSyn}

    Linkage synthesis is a classical kinematic design task that involves constructing a linkage mechanism to transfer an input motion (typically a rotation input from a motor) to an output motion that satisfies a set of specified characteristics~\cite{acharyya2009performance}. Two typical linkage synthesis problems include rigid body guidance and path generation~\cite{antonsson2001formal}. The goal of rigid body guidance is to lead a rigid body past a series of given positions and orientations~\cite{zhixing2002study}. Several recent works have addressed this problem in designing mechanical automata, including planar mechanical characters~\cite{coros2013computational}, multi-pose mechanical objects~\cite{nishida2019multi}, and drawing machines~\cite{roussel2017spirou}. Instead of focusing on rigid body motion, path generation aims to generate a mechanical linkage that enables the end-effector point to move along a desired trajectory. Due to its simplicity and ease of fabrication, the 1-DoF planar linkage is the most widely used mechanism for generating 2D paths. Some researchers have developed computational tools to address the 2D path generation problem, such as mechanical character design~\cite{coros2013computational, thomaszewski2014computational} and interactive editing methods~\cite{bacher2015linkedit}. Additionally, they have explored walking machines~\cite{bharaj2015computational}. Cheng et al.~\cite{cheng2021spatial} leverage the 3D cam-follower mechanism to generate a 2D path on a planar surface and a cylindrical or spherical surface.

    Furthermore, much effort has been made to generate a 3D path~\cite{nollexa1975linkage, cervantes2011synthesis, lin2018geometric}. One of the typical solutions is to use 1-DoF spatial linkages, whose links can move in 3D space and be driven by a single actuator, such as the RCCC mechanism~\cite{bai2022exact} and the RRSS mechanism~\cite{liu2020synthesis}. Similar to their 2D counterparts, these designs can approximate a desired 3D path but still have limited engineering application cases. While little literature addresses the path generation problem with the overconstrained linkage family, it has been demonstrated for its engineering potential in robotic applications~\cite{Gu2022OverconstrainedCoaxial} and energy efficiency in robotic limb design~\cite{gu2023computational}. Compared with 1-DoF spatial linkages, Cheng et al.~\cite{cheng2022exact} combine the 3D cam-follower with a spatial linkage mechanism for exact path generation. However, the proposed mechanism has limited application in the robotics field due to its higher number of pairs and lack of stiffness. In contrast, our overconstrained mechanism consists of only revolute joints, which are convenient to maintain and assemble and capable of carrying a relatively high payload. As its planar counterpart (planar four-bar linkage) has been widely used in modern machinery~\cite{antonsson2001formal}, the overconstrained linkage should have sufficient potential for advanced robotic design~\cite{Gu2022OverconstrainedCoaxial}.
    
\subsection{On Geometric Generation}
\label{sec:Intro-GeoGen}

    One of the most widely studied geometric generation problems is the topology optimization problem, aiming to optimize the geometric shape of static objects under dynamic constraints~\cite{sigmund2013topology}. There is relatively little literature on optimizing the geometric shapes of dynamic objects under kinematic constraints, such as designing self-collision-free structures. In the graphics community, recent work has focused on the design of transformable objects~\cite{zhou2014boxelization}, which are typically realized via 3D printing. For example, Yuan et al.~\cite{yuan2018computational} propose a computational approach to generate transformable robots that shape-shift into different forms. However, the design of these objects extends beyond linkages and typically only concerns the initial and final states~\cite{huang2016making}. The geometric design of the 2D linkage can be resolved by straightforwardly offsetting each component along the directional normal to its motion plane~\cite{coros2013computational}.

    Regarding spatial linkages, it becomes a challenge to design collision-free geometries due to the increased dimensions~\cite{anvari2018collision}. Li et al.~\cite{li2020invertible} leverage the boolean operation to trim the link design for generation collision-free linkage, as well as imply the conjecture of the existence of collision-free spatial linkages over the complete motion circle. However, the resulting link design is defined by the swept volume and can not be directly assembled for a practical linkage mechanism. Zhang et al.~\cite{zheng2016deployable} transfer the link bar into the deformable chain to obtain the collision-free deployable 3D structure. At the same time, their linkages do not aim at invertible motion and are not directly applicable to robotic structure design. In addition, their method employs a gradient-free approach that requires a relatively longer computing time than our formulation. Therefore, unlike the existing literature, our work aims to generate self-collision-free robotic limb design by a gradient-based formulation for engineering purposes.
    
\subsection{Proposed Method and Contributions}

    This study presents a computational design method for energy-efficiently realizing user-specified spatial motion via the self-collision-free overconstrained linkage driven by a single actuator. We proposed a generalized method for self-collision-avoidance design by formulating the geometric optimization problem with continuous constraints, which is solvable by non-linear programming problem solvers. Then, we formulated the motion design problem with 1-DoF overconstrained linkages and optimized the design parameters to obtain energy-efficient target trajectories. Finally, we conducted several experiments to validate the proposed method and designs, including kinematic and dynamic evaluations, mechanical character demonstrations, and a hexapod prototype robot. We found that the overconstrained robotic limb design shows the potential to generate energy-efficient motion in specific tasks. The contributions of this study are listed as follows:
    \begin{itemize}
         \item Proposed a generalized gradient-based design approach for linkage-based robotic limbs with self-collision-free motions.
        \item Developed a computational design framework by optimizing the similarity and energy-related metrics for 1-DoF overconstrained robotic limbs with desired trajectories.
        \item Validated the kinematic and dynamic performance of the resulting designs and demonstrated the superior energy efficiency of overconstrained limbs with a hexapod walker.
    \end{itemize}

    The remainder of this study, Section~\ref{sec:Method}, presents the collision-free design method and optimization problem formulation for overconstrained motion generation. Validation with hardware experiments and further discussion are enclosed in Sections~\ref{sec:Result} and~\ref{sec:Discuss}. Section~\ref{sec:Final} presents the conclusion, limitations, and future work, which conclude this work.

\section{Method}
\label{sec:Method}

\subsection{Generalization of Collision-Free Design for Robotic Limbs}
\label{sec:Method_CollideFree}

\subsubsection{Parameterized Link Bar}

\begin{figure}[ht]
        \centering
        \includegraphics[width=0.8\columnwidth]{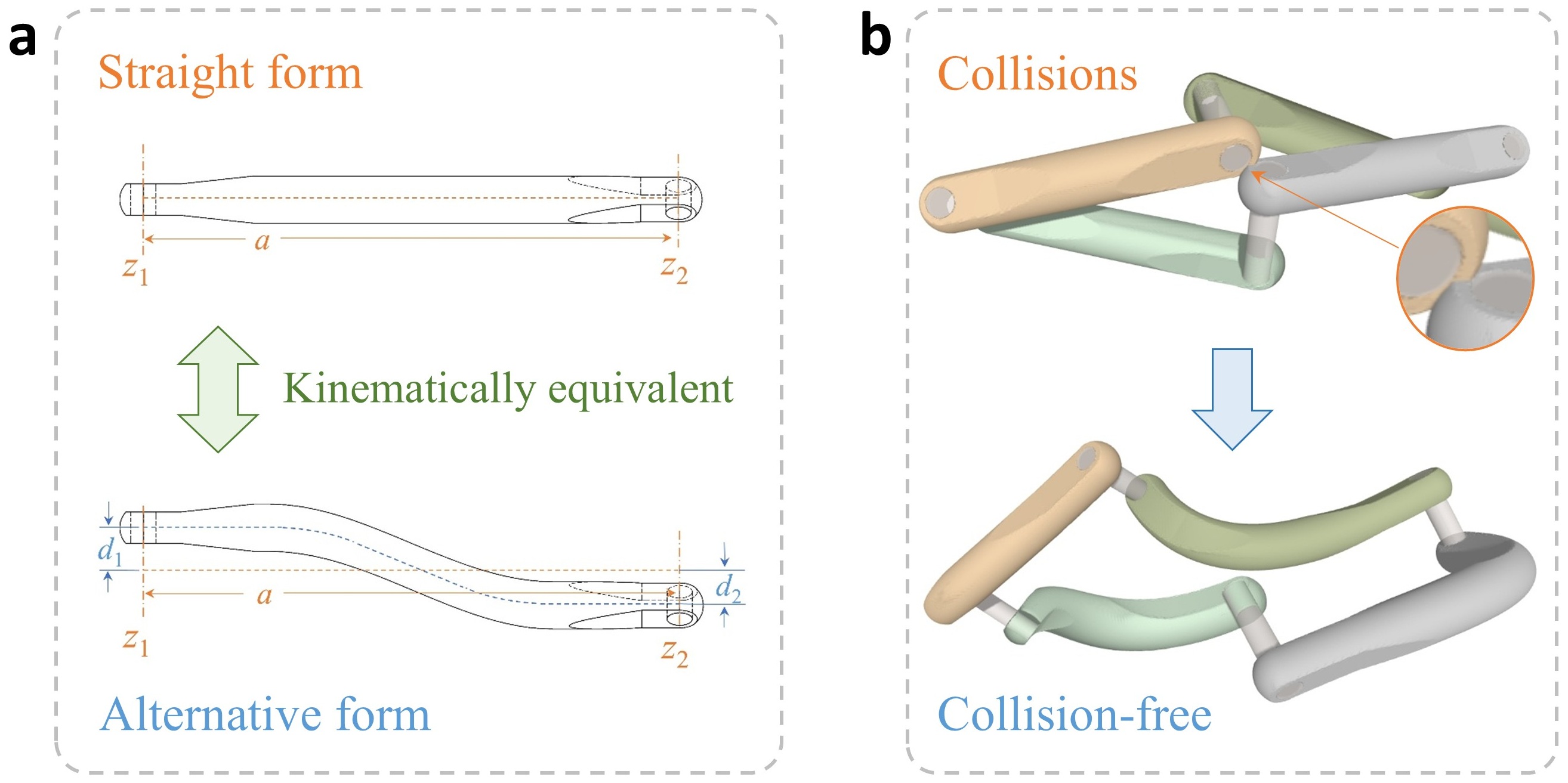}
        \caption{Deform the link bar to avoid collisions.}
        \label{fig:Method_CollideFree_CollisionExplained}
    \end{figure}

    The spatial linkage inevitably collides with itself when the actuator implements a wide range of rotation. In this section, we fix the linkage's kinematic parameters but modify its geometrical shape to avoid collisions. To ensure the engineering feasibility, our intuition is to design an alternative form of linkage by adjusting the geometric offset along the rotational axis. The resulting link is a curved bar with an equivalent cross-sectional area for rod radius and joint size (see Fig.~\ref{fig:Method_CollideFree_CollisionExplained}). 
    Without loss of generality, we parameterize the central axis line of each link by polynomials of degree $n$ as
    \begin{equation}
        \Psi _{i}(s)=a_{0}+a_{1}s+\cdots +a_{n}s^{n},
        \label{eq:link_representation}
    \end{equation}
    where $s\in [0,1]$ is the parameter that can be used to describe any position of a point located on the central axis of the link, and the polynomial coefficients $a_{i}$ describe the geometry of the link. 

\subsubsection{Continuous collision detection}

\begin{figure}[ht]
        \centering
        \includegraphics[width=0.8\columnwidth]{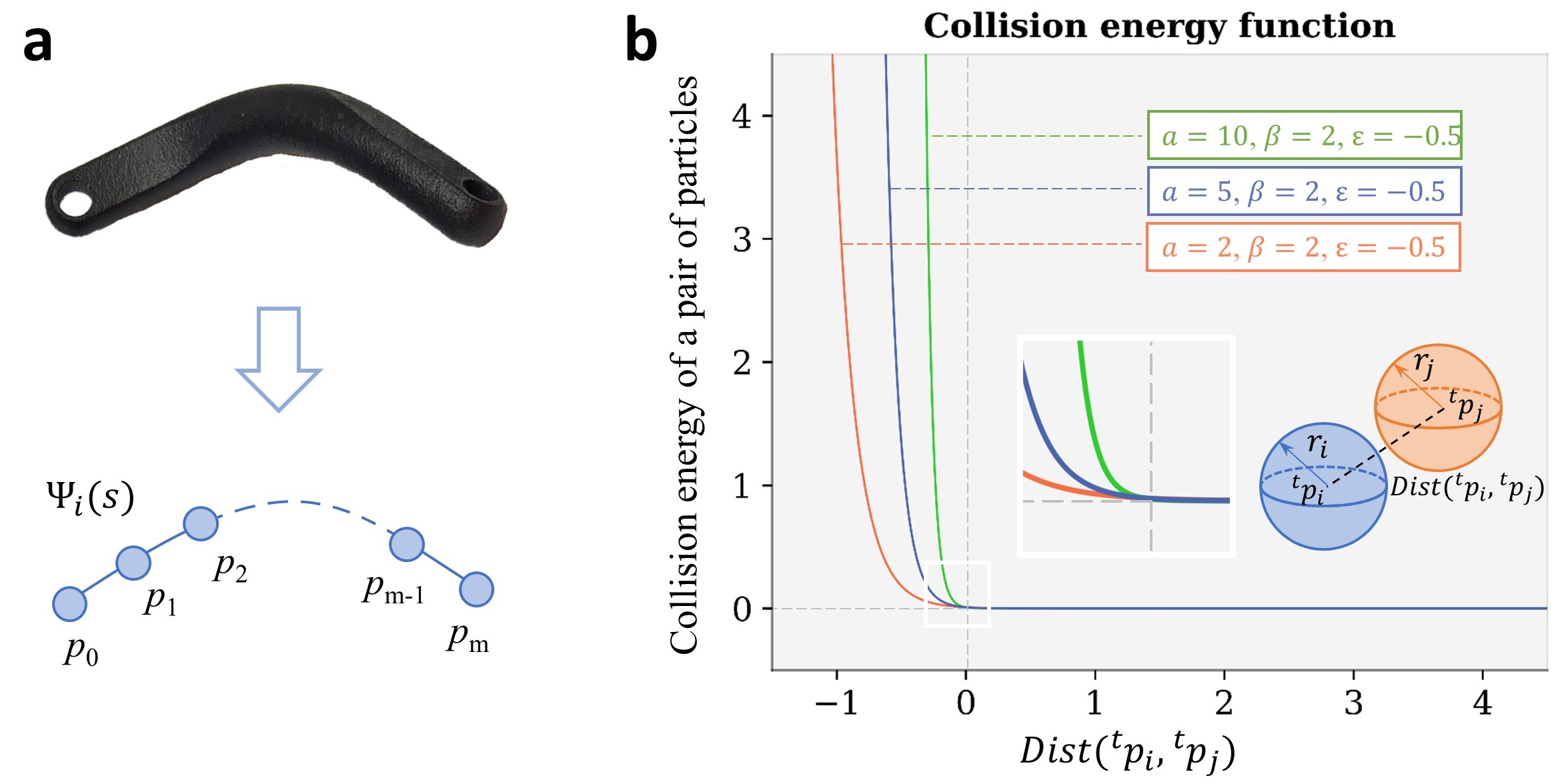}
        \caption{(a) Represent the linkage using a chain of particles. (b) Continuous collision energy term. $Dist(^{t}p_{i}, ^{t}p_{j})$ describes the distance between particle $i$ and $j$ at time frame $t$,}
        \label{fig:Method-ParaAndFun}
    \end{figure}

    By sampling the parameter $s$ uniformly, we can discretize each bar into a chain of particles; see Fig.~\ref{fig:Method-ParaAndFun}. 
    And the radius of particles is equivalent to the bar thickness. Therefore, the optimization variables are the positions of each particle. Conceptually, we can regard each link as an elastic rod with specific kinematic parameters, and colliding with other links would deform their geometry plastically. Due to kinematic constraints, the start and end particles must always be located on the rotational axis. Borrowing concepts from continuum mechanics, we define the collision energy term for particles of each chain to constrain the deformation process. Specifically, the function $f(x)$ should be defined such that $f(x)\rightarrow 0$ as $x>0$, ${f}'(x)\rightarrow 0$ as $x>0$, and $f(x)\rightarrow \infty $ as $x<0$. Instead of the piece-wise function in previous work~\cite{kaldor2008simulating}, our approach is to create a smooth, continuous function that approximates the energy term~\cite{rakita2018relaxedik}. The function is defined as follows:
    \begin{equation}
        f_{m,n}(^{t}p_{i}, ^{t}p_{j})=\left [ 1+\tanh\left ( \alpha Dist(^{t}p_{i}, ^{t}p_{i})+\beta \right ) \right ]^{-1}+\varepsilon,
    \end{equation}
    and
    \begin{equation}
        Dist(^{t}p_{i}, ^{t}p_{j})=\left \| R_{m}(t)^{0}p_{i}-R_{n}(t)^{0}p_{j} \right \|_{2}-r_{i}-r_{j},
    \end{equation}
    where $^{t}p_{i}$ and $^{t}p_{j}$ are the positions of particle $i$ and $j$ at time frame $t$, $R_{k}(t)\in SE(3)$ (Special Euclidean group in 3 dimensions) is the homogeneous transformation matrix of linkage $k$'s forward kinematics at time frame $t$,  $r_{i}$ and $r_{j}$ are the radii of particle $i$ and $j$, respectively. $Dist(^{t}p_{i}, ^{t}p_{j})$ describes the distance between particle $i$ and $j$ at given time frame $t$, which is less than zero when the collision occurs. And $\alpha$, $\beta$, and $\varepsilon$ are constant model parameters. In detail, the $\alpha$ value defines the gradient amplitude. In contrast, a large $\alpha$ would improve collision detection accuracy but increase the risk of exploding gradient; see Fig.~\ref{fig:Method-ParaAndFun}(b). In our prototype solver, we implement a set of values $\alpha=5$, $\beta=2$, and $\varepsilon=0.5$. Essentially, this function is consistent with prior collision-avoidance techniques, which is close to zero when there is no collision and high otherwise~\cite{khatib1986real, nakamura1990advanced}. Then, we sum up the collision energy term between all pairs of links at every discrete time frame $t$ over the total time duration as the total collision energy:
    \begin{equation}
        E_{col}(N,K,T)=\sum_{(m,n)}^{N}\sum_{(i,j)}^{K}\sum_{t=0}^{T}f_{m,n}(^{t}p_{i}, ^{t}p_{j}),
    \end{equation}
    where $N$ represents the set of all the link pairs $m$ and $n$, $K$ represents the set of all the particle pairs $i$ and $j$ between link $m$ and $n$, and $T$ is the total time duration. Note that the above formulation is smooth and differentiable, allowing for the calculation of gradients for further optimization.

\subsubsection{Geometric optimization problem}
\label{sec:geom_opt_problem}

    We formulate the geometry design problem as a continuous optimization problem with non-linear constraints. In this stage, the kinematic parameters and link thickness are fixed. Also, the forward kinematics of each link has been explicitly derived during the whole period $T$ in section \ref{subsec:kin_model}. Without loss of generality, we assume the actuator rotates with a constant angular speed $\omega$ over the complete motion cycle when simulating the mechanism's kinematics. Since each link can be parametrically described in the form of Eq.~\eqref{eq:link_representation}, the optimization variables are the list of polynomial coefficients that describe the geometry of links, $\hat{\Omega} =\left [ \Psi _1,\Psi _2,\cdots \Psi _n \right ]$. In this study, we implement the scissor-like revolute joint as the central motion pair unit, while one can also introduce almost any of the kinematic pairs with explicit kinematics. For fabricating feasibility, we ensure the resulting design has identical revolute joint dimensions, determined by their overall size. Note that collisions between joints and links should also be considered. Since we assume the mechanism configuration is given, we can treat the revolute joint as non-deformable straight link elements with specified starting and ending points defined by the endpoints of other links. Therefore, the geometric optimization problem can be formulated to find the feasible solution $\hat{\Omega} _{f}$ that satisfies the following constraints within the total duration $T$.

    As listed below, we formulate diverse constraints related to collision avoidance, kinematics, and fabrication.
    \begin{enumerate}
        \item \textit{Collision-free motion}. To ensure the collision-free motion, we require the total collision energy term to be no larger than a threshold $E_{th}$ during the whole time duration $T$.
        \begin{equation}
            E_{col}(N,K,T)< E_{th},
        \end{equation}
        where $E_{th}=0.1$ in our prototype experiments.
        
        \item \textit{Link length}. The lengths of each link should be optimized within a window about the kinematic design parameter $L_{k,0}$ defined by:
        \begin{equation}
            L_{k,0}\leq \sum _{i=0}^{m}\left \| \Psi_{k}(s_{i+1})-\Psi_{k}(s_{i}) \right \|_{2}\leq  \xi _{L} \cdot L_{k,0},
        \end{equation}
        where $m$ is the particle number of each link $k$, $s_{i}\in[0,1]$ is the sampling parameter, and $\xi _{L}$ is a coefficient typically set as 1.3 in our experiments.
        
        \item \textit{Link curvature}. To ensure physical feasibility, we restrict the discretized curvature at particle $i$ that should always be no larger than the threshold value $\kappa _u$:
        \begin{equation}
            \kappa_{i,k} =\frac{2\vec{e}_{i+1}\times\vec{e}_{i} }{\left \| \vec{e}_{i+1} \right \|_{2}\left \| \vec{e}_{i} \right \|_{2}+\vec{e}_{i+1}\cdot \vec{e}_{i}} \leq \kappa _u, i\in [0,m],
        \end{equation}
        where $ \kappa_{i,k}$ is the discrete curvature at vertex $k$~\cite{bergou2010discrete}, $\vec{e}_{i}$ is the particle vector $i$ defined by particle $\Psi(s_{i})$ and $\Psi(s_{i+1})$. The curvature constraint prevents the link from generating excessive deformation while avoiding collisions.
        
        \item \textit{Rotation axis error}. Each link's start and end points should be located at the rotation axis to ensure the validity of its kinematics. We compute the distance of the link's endpoint to the rotation axis $\vec{z}_{j}$:
        \begin{equation}
            \frac{\left \| \vec{v}_{j,k}\times \vec{z}_{j} \right \|_{2}}{\left \|  \vec{z}_{j} \right \|_{2}^{2}}< \xi_{c} ,  1\leq k\leq N, 1\leq j\leq N, 
        \end{equation}
        in which $\vec{v}_{j,k}$ the vector is defined by the link $k$'s endpoint $\Psi(s_{m})$ and any point on the axis $j$, $z_{j}$ is the direction vector of the rotation axis. When the link's endpoint is located on the rotation axis, the above equation should be equal to zero. And $\xi_{c}$ is the allowed axis distance error (unit: millimeter), set as 0.01 in our experiments.
        
        \item \textit{Revolute joint thickness}. The revolute joint thickness of any pair of links should be equivalent to a constant value $D_{j}$ to unify the hinge design:
        \begin{equation}
            \left \| \Psi_{k}(s_{m})-\Psi_{k+1}(s_{0}) \right \|_{2}=D_{j},
         \end{equation}
         in which $s_{m}$ is effectively equal to 1 and $s_{0}=0$. $D_{j}$ is the distance from the endpoint of link $k$ to the start point of link $k+1$. And the final joint thickness should be equal to $2r+D_{j}$, where $r$ is the radius of the joint part. The joint center distance $D_{j}$ must be larger than the link particle's diameter. A smaller $D_{j}$ would result in a more compact design, while the optimization problem might be harder to converge since the links are more likely to collide.
         
         \item \textit{Perpendicular to rotation axis}. To ensure the assembly performance and kinematic feasibility, the start and end line segments of each link should be perpendicular to the corresponding rotation axis:
         \begin{equation}
            \vec{e}_{0,k}\cdot \vec{z}_{j}=0,\:  \vec{e}_{m-1,k}\cdot \vec{z}_{j+1}=0,
         \end{equation}
         where $\vec{e}_{0,k}$ is the start line segment of link $k$, $\vec{e}_{m-1,k}$ is the end line segment of link $k$, and $z_{j}$ is the direction vector of the rotation axis.
    \end{enumerate}

    Note that the designer can also formulate the link length or curvature as the objective function to resolve this problem; however, this may require more computing time to converge to the optimum design. In this prototype study, we primarily focus on finding a feasible solution with relatively low computational time consumption. Our computational design methodology can generate physically feasible designs for collision-free robotic limbs or provide instructive insights into constructing the geometric shape for a system with complicated kinematics.
    \begin{figure}[t]
        \centering
        \includegraphics[width=0.9\linewidth]{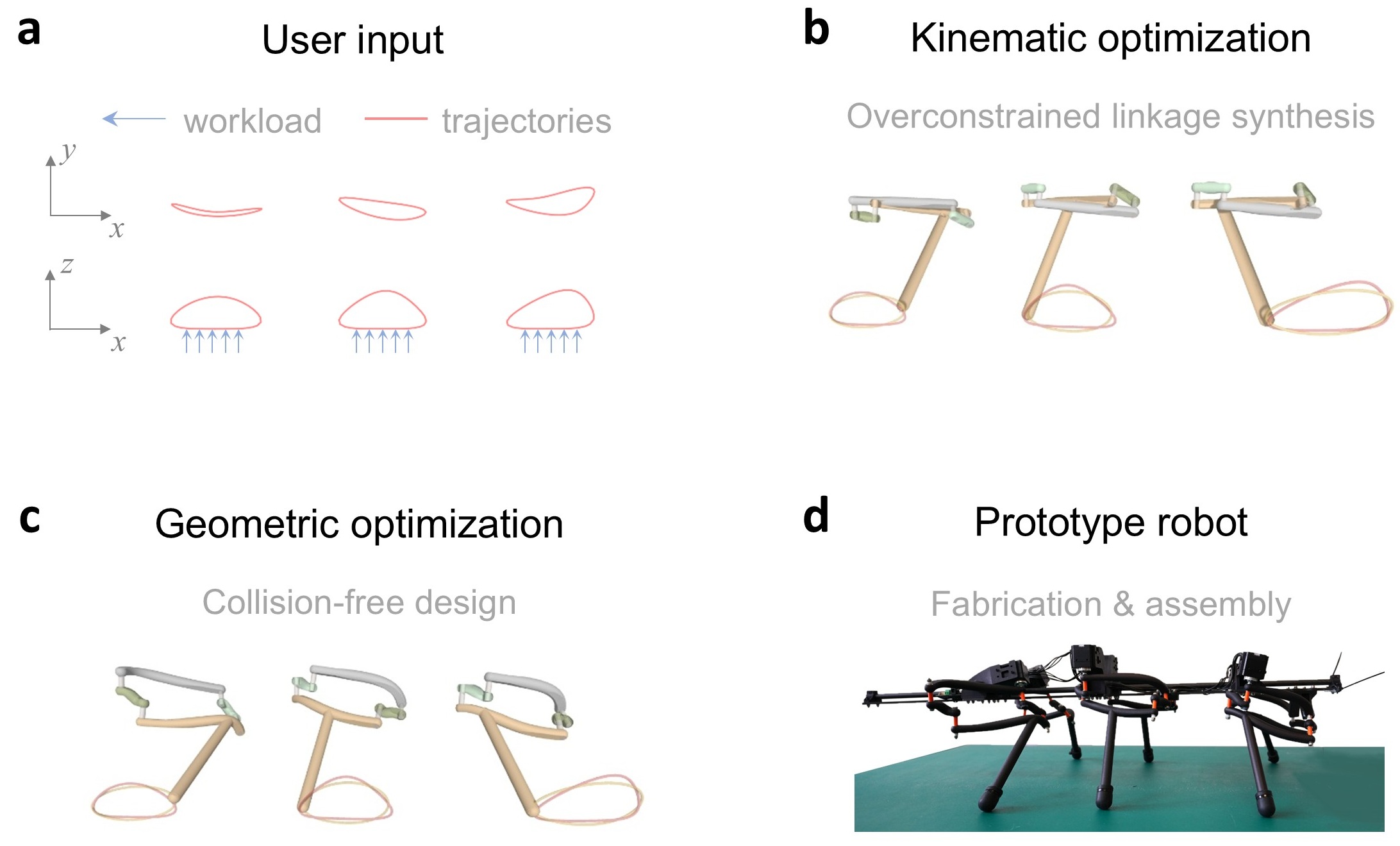}
        \caption{Overview of our approach. (a) Target motions are provided, including the spatial curve, time interval points, and workload; (b) optimized overconstrained robotic limb generated by resolving the parameter optimization; (c) collision-free design generated by geometric optimization stage; and (d) fabricate the final design by 3D printing and assemble with other parts.}
        \label{fig:Method-Overview}
    \end{figure}

\subsubsection{Initialization}
    
    Since our geometric design problem is highly non-linear and non-convex, a naively generated initial design vector is likely to be far away from the feasible area. It may lead to slow convergence or overall failure of the optimization (exceeding the maximum number of iterations). Hence, it is essential to find a suitable initial guess before enforcing the geometric constraints (Eqs. 5-10). The initialization stage is composed of two parts: the constrained random parameter generation phase and the evaluation phase. First, we randomly initialize the start and end points of each link, which are located at the rotation axis and satisfy the revolute joint thickness constraint. Then, we define the cubic polynomial coefficients of the initial link design's central axis line by utilizing its position values (start and end points) and derivative at the end-points (the central line should be perpendicular to the rotation axis). This intuitive method is also known as the \textit{Hermite} parameterization. After the generation phase, we evaluate the design candidates by accumulating their total collision energy, $E_{col}$, over the entire time duration, $T$. The design with the highest score (lowest collision energy) is chosen as the candidate to continue to the further geometric optimization stage. In practice, the designers are suggested to start the optimization from different initial design vectors and select the one with the best structure. We can also initialize our method using the warm-starting technique. For example, the designer can easily obtain a set of solutions under relaxed constraints. Then, taking these solutions as input and restricting the constraints (adding the rod radius or reducing the joint thickness), the algorithm may rapidly converge to a nearby solution, which would decrease the optimization time for a challenging constraint setup.

\subsubsection{Optimization Solver}
    
    Our optimization problem is challenging to resolve due to many constraints, some of which are non-convex, such as the collision energy term. The collision constraint is practically enforced every 0.025 \unit{\second}, which is determined by the angular velocity of the actuator, while other constraints are checked at the initial state. Additionally, it is crucial to set up proper sampling of particle numbers concerning their radius. A larger number of particles and a shorter time interval for checking may slightly improve the accuracy of collision detection. Still, it would significantly increase the complexity of collision constraints and the possible computation time. In our experiments, we sample each link with 20 particles and each joint with five particles. To address the non-linear optimization problem, we implement the Interior Point Method solver with a Python wrapper (Ipopt)~\cite{wachter2006implementation}. We set the maximum number of iterations for the solver to 1500, and the average solving time is nearly 10 minutes. These values vary depending on the problem definition, including the joint center distance, number of particles, and other parameters. All the results are computed using an Intel Core i7-13700F 2.1 GHz CPU. To speed up the optimization, the Jacobians and Hessians of constraints are calculated analytically by CasADi~\cite{andersson2019casadi} and provided to the solver, which is crucial for performance. Another factor that reduces the computing time is that we do not use a cost function, as formulated in section~\ref{sec:geom_opt_problem}. This also reduces the number of tuning parameters, as the relative importance of constraints does not need to be quantified, since all must be satisfied for a design to be physically feasible.

\subsection{The Mechanism Synthesis Problem with Overconstrained Linkages}
\label{sec:Method-Formulation}

\subsubsection{Overconstrained Linkage Synthesis}
\label{sec:Problem}
    
    Our goal is to design an overconstrained linkage mechanism that can transfer the rotational motion of a single actuator to the periodic spatial trajectories specified by the user, as shown in Fig.~\ref{fig:Method-Overview}. In the following, we define the overconstrained linkage and describe the user-specified motions.

    \textit{Overconstrained Mechanism.} Overconstrained mechanisms in this study refer to the spatial linkage family with a single closed loop and revolute-only joints. According to Chebychev–Grübler–Kutzbach's mobility criterion~\cite{gogu2005chebychev}, this class of mechanisms is classified as overconstrained due to their mobility from the specifically designed geometry. Therefore, movable $nR$ loops with $n < 7$ are overconstrained or paradoxical. The only 4$R$ overconstrained mechanism is named the Bennett linkage, which is also one of the simplest revolute-only spatial linkages with the least number of links. This study primarily utilizes the Bennett linkage to generate rational spatial motion with a single actuator. The motion of the Bennett linkage can be represented by a conic section of the Study quadric~\cite{hamann2011line}. Additionally, by merging the Bennett linkages or developing symmetric constraints, we can formulate various designs of overconstrained 5R or 6R linkages. The increased number of links may generate a more complex rational spatial curve, which will be addressed in the discussion section.

    \textit{User Input.} The user input involves $n$ smooth and closed parametric curves $C_{i}\left ( s \right )$, $s\in [0,1], 1\leq i\leq n$, which are properly arranged in the 3D space. Each curve also represents the desired trajectory of the end-effector, as shown in Fig.~\ref{fig:Method-Topology}. Due to the finite number of design parameters, we can only approximate the given curves. Therefore, we restrict the user input to the type of spatial B-spline curves with $k$ control points. The users can directly drag the control points to generate their desired spatial trajectories.

    On the other hand, users can also provide the collected spatial trajectory data points as input. Note that the Bennett linkages can only represent spatial curves with relatively simple shapes. When performing more complex spatial curves, the overconstrained mechanism with more links might be able to approximate them properly. Additionally, users can implement temporal control of motion by inserting $m$ key points on each curve. Also, the time interval $T_{i}^{p}$ of each of the two pairs of key points should be specified, where $\sum T_{i}^{p}=1$. A relatively constant linear velocity would be kept between each pair of key points. By adjusting the target trajectories and timings, the user can design the automaton with realistic functional behavior, such as boating or walking.

    \textit{Design Requirements.} Given the user input, our computational design framework should generate the optimized linkage mechanism to satisfy the following high-level requirements:
    \begin{enumerate}
        \item \textit{Motion trajectory and timing}: The motion trajectories of the linkages' end-effectors should approximate the input curves as much as possible, exactly passing the defined critical points at a given time.
        \item \textit{Lightweight and simple mechanism}: The overconstrained linkage mechanism should be lightweight to reduce the driven torque and have as few mechanical parts as possible.
        \item \textit{Energy-efficient motion}: The motion of an overconstrained linkage should be energy-efficient for a given workload to adopt a wide range of application scenarios.
        \item \textit{Collision-free motion}: No collision is allowed during the full-cycle motion among these spatial linkages.
        \item \textit{Singularity-free and smooth motion}: The motion should avoid the singularities of the spatial overconstrained linkages and maintain low acceleration.
    \end{enumerate}

    In the following section, we first define the topology of our linkage mechanisms for spatial trajectory generation, which have the fewest links and kinematic pairs. We formulate the kinematics and dynamics of the linkage mechanism analytically with a generalized method. Building on the 3D path similarity metric and mechanism modeling as the foundation, we address the optimization problem of the linkage design parameters for a given input and objectives. After optimizing the link parameters, we apply the geometric design method to generate a collision-free geometry for the spatial overconstrained mechanism, ensuring it meets the remaining high-level requirements.

\subsubsection{Topology Modeling and Parameter Space}

\begin{figure}[ht]
        \centering
        \includegraphics[width=0.9\columnwidth]{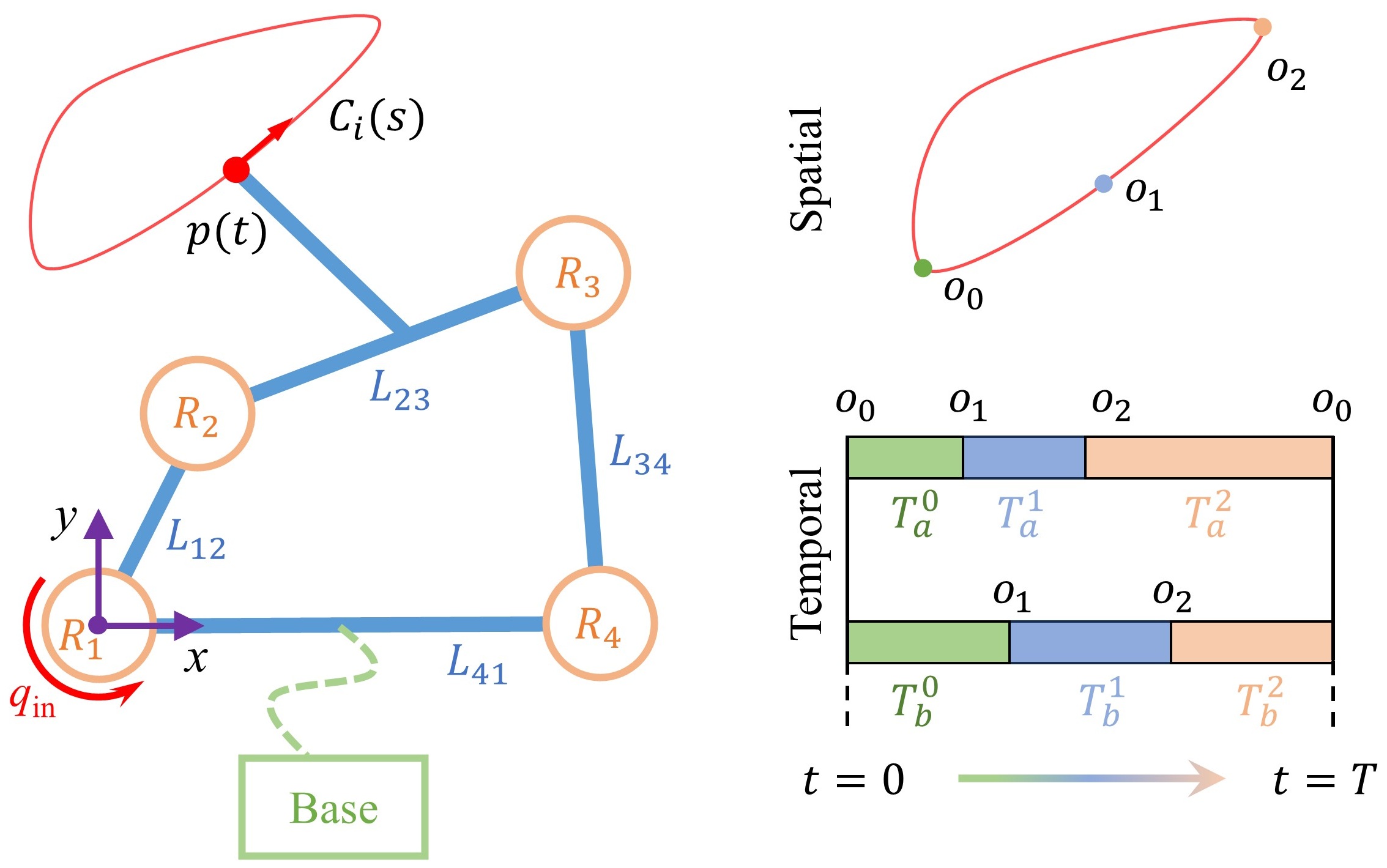}
        \caption{The example topology configuration modeled in this study. The input of our approach includes spatial curves, key points for temporal control, and workload conditions.}
        \label{fig:Method-Topology}
    \end{figure}

    Given an overconstrained linkage, we define the local coordinate system in the following way:  the origin is located at the middle point of the rotation axis between the base link and the active one, the rotation axis also defines the $z$-axis in clockwise, and the $x$-axis is along the direction of the base link (see Fig.~\ref{fig:Method-Topology}). 
    The local coordinate system defines the kinematic and dynamic modeling variables without additional specification. The topology configuration of our design is relatively straightforward, consisting only of revolute joints and twisted links with zero offset. Kinematically, each link part consists of only two parameters: link length and twist angle. The topology of overconstrained linkages consists of $n$ spatial links ($n<7$) connected by revolute joints, forming a single closed loop and movable with only one degree of freedom. Taking the only 4$R$ case (Bennett linkage) as an example, it must satisfy the following geometric constraints~\cite{Baker1979TheBennett}: 
    \begin{equation}
        a_{12}=a_{34}=a,\:a_{23}=a_{41}=b,
    \end{equation}
    \begin{equation}
        \alpha_{12}=\alpha_{34}=\alpha,\: \alpha_{23}=\alpha_{41}=\beta,
    \end{equation}
    \begin{equation}
        \frac{\sin\alpha }{a}=\frac{\sin\beta }{b}=\gamma,
    \end{equation}
    \begin{equation}
        R_1 = R_2 = R_3 = R_4 = 0,
    \end{equation}
    where $a_{ij}$ and $\alpha_{ij}$ represent the link length and twist angle of link $L_{ij}$, respectively. $R_i$ is the offset of link $L_{ij}$ at rotation joint $i$. $\gamma$ is the Bennett radio between the link length $a$ and the sine of twist angles $\alpha$. As illustrated in Fig.~\ref{fig:Method-Topology}, by choosing link $L_{41}$ as the base link that supports the entire mechanism, the spatial linkage chain is movable with a single degree of freedom by actuating link $L_{12}$ (the active link) along axis 1. Link $L_{23}$ and $L_{34}$ are passive links. To generate spatial motion, the end-effector must be placed on the passive link $L_{23}$, as the active link $L_{12}$ and the passive link $L_{34}$ can only output a 1-DoF rotating motion. Therefore, the geometry of the end-effector is modeled as a bar extended from the link $L_{23}$. The endpoint on the end-effector for generating the spatial motion is defined as the end-effector point, denoted as $p$ (see Fig.~\ref{fig:Method-Topology}). The initial position of the end-effector determines the parameters of the end-effector point $p_{0}=[x_{p,0},\: y_{p,0},\: z_{p,0}]$. Hence, the kinematic design parameters $\mathcal{X}_{i}=[\phi_{i},\: p_{0},\: ^{i}T_{w}]$ can be represented by a set of each link's parameters $\phi_{i}$, end-effector parameters $p_{0}$, and the homogeneous transform matrix from the world frame to the local one $^{i}T_{w}$. Note that the shape of the generated spatial curve is only determined by the first two terms, and the modification of the transform matrix only results in the translation and orientation of the output curve. The transform matrix $^{i}T_{w}$ must be considered when it involves dynamic analysis and collision avoidance. Once the kinematic design parameters are determined, we can obtain the list of actuators' input angles $q$ based on the given key points, which serve as additional timing control parameters.

\subsubsection{Kinematic modeling}
\label{subsec:kin_model}

    In this section, we first present our approach to modeling the kinematics of the 4$R$ Bennett linkage, considering joint $R_{1}$ as the active joint and link $L_{23}$ as the coupler. Since the topology of our spatial linkage system is a closed loop, the kinematic chain $R_{1}\rightarrow R_{2}$ is kinematically equivalent to the one $R_{4}\rightarrow R_{3}$. Therefore, we formulate the forward kinematics of the overconstrained linkage using the product of the exponential formula via one of the kinematic chains as:
    \begin{equation}
        p(t)=e^{[\mathcal{S}_1]\theta_1} e^{[\mathcal{S}_2]\theta_2} P_0,
    \end{equation}
    \begin{equation}
        \theta_{1}=q(t), \: \theta_{2}=K(\theta_{1}),
    \end{equation}
    where $\mathcal{S}_{i}\in \mathbb{R}^{6}$ is the initial velocity screw of joint $R_{i}$ $(i=1,2,3,4)$, $P_{0}$ is the initial position of the end-effector point, and $\theta_{i}$ is the corresponding joint angle. Note that $K$ represents the mapping relationship from the active joint angle $\theta_{1}$ to the passive one $\theta_{2}$, which can be obtained from the closure equations of overconstrained linkages. Given the actuator input angle $q(t)$, $t\in [0,T]$, the active joint angle $\theta_{1}$ is equivalent to the actuator input. The above formulation can also be extended to the kinematic analysis of 5$R$ and 6$R$ cases. Then, the velocity screws $\mathcal V$ of the end-effector $p$ can also be obtained as the following:
    \begin{equation} \label{eq:jacobian}
        \mathcal V=J_{\mathcal{S}}(\theta)\dot{\theta }=\begin{bmatrix}
        J_{\mathcal{S}_{1}}&J_{\mathcal{S}_{2}}
        \end{bmatrix}\begin{bmatrix}
        \frac{\mathrm{d} \theta_{1}}{\mathrm{d} q} &
        \frac{\mathrm{d} \theta_{2}}{\mathrm{d} q}
        \end{bmatrix}^T\dot{q}=J(q)\dot{q},
    \end{equation}
    in which $J_{\mathcal{S}_i}=Ad_{e^{[\mathcal S_1]\theta_1}\cdots e^{[\mathcal{S}_{i-1}]\theta_{i-1}}}(\mathcal S_i) \in \mathbb{R}^{6}$ is the screw vector for joint $i$ expressed in space frame coordinates, with the joint values $\theta$. Then, we can also derive the analytical forward kinematic Jacobian matrix $J_{a}(q) \in \mathbb{R}^{3\times1}$ from the above equation, which maps the actuator's angular velocity into the end-effector's linear velocity vector $\dot{p}=[\dot{x},\dot{y},\dot{z}]^T$. And the condition number $\kappa (J_{a}J_{a}^T)$ describes the distance from the linkage state to its singular configuration~\cite{Merlet2006Jacobian}. Based on the kinematic formulation, the condition number $\kappa$ can evaluate the smoothness of the linkage motion.

\subsubsection{Modified spatial curve similarity metric}

    Once the kinematic parameters are given, based on the FK formulation, we can obtain the spatial path of the end-effector, which is represented by a sequence of scattered points $C=[p_{1},p_{2}, ..., p_{n}]$, $p_{i} \in \mathbb{R}^{3}$. Similarly, the user's input can be transferred into the same form. To determine the design parameters of the overconstrained mechanism that best approximates a user's input curve, we must measure the similarity between two spatial curves. While one of the possible solutions is to represent a 3D curve as two 2D projections, then leverage the well-studied similarity metric for 2D curves, such as the Hausdorff distance and $\rm Fr\acute{e}chet \: distance$~\cite{alt1995computing}. However, the projection in the 2D plane is not invariant to the rotation of the spatial curve in 3D space. Therefore, we proposed the modified spatial curve similarity metric based on the curvature feature inspired by the method used in~\cite{rodriguez20043}. The first step of our approach is to transfer the 3D curve into a set of uniform discrete points. Then, we can represent the spatial curve with a sequence of 2-element curvature-based feature vectors as:
    \begin{equation}
        C=\begin{bmatrix}
        \begin{pmatrix}
        \gamma _1\\
        \delta _1
        \end{pmatrix},& \begin{pmatrix}
        \gamma _2\\
        \delta _2
        \end{pmatrix},& ... ,& \begin{pmatrix}
        \gamma _n\\
        \delta _n
        \end{pmatrix}
        \end{bmatrix},
    \end{equation}
    where $\gamma_{i}$ is the angle between the line tangents $\overline{p_{i-1}p_{i}}$ and $\overline{p_{i}p_{i+1}}$, and the $\delta_{i}$ is the angle between the binormal vectors of two planes, one is defined by the line segments $\overline{p_{i-1}p_{i}}$ and $\overline{p_{i}p_{i+1}}$. The other plane is defined by the line segments $\overline{p_{i}p_{i+1}}$ and $\overline{p_{i+1}p_{i+2}}$. Note that each element $\gamma_{i}$ and $\delta_{i}$ is in the range of $[0,2\pi]$. To this end, we use the minimum distance values obtained by selecting the starting index $k$ of one of the 3D curves by computing the $L_{1}$ norm of the curvature difference and define the similarity metric as:
    \begin{equation} \label{eq:similarity}
        E_{s}(A,B)=\frac{1}{2\pi n} \min_{k} \sum_{i=0}^{n} \left( \left| \gamma_{A,k+i}-\gamma_{B,i} \right|+\left| \delta_{A,k+i}-\delta_{B,i} \right| \right).
    \end{equation}
    Note that the proposed similarity metric is normalized into the $[0,1]$ range and invariant to either translation, rotation, scaling, or the starting point on a closed spatial curve. Therefore, given two spatial curves, if they have similar shapes, the similarity metric would be close to zero, and vice versa. The proposed metric leverages the curvature feature of spatial curves and is suitable for quantifying the similarity between any two sets of closed spatial curves. We can accelerate the computing process using the dynamic programming technique. However, it is noted that our approach requires the curves to have the same number of discrete points. If this is not the case, we will resample the coarser one.

    Besides the similarity metric in the objective function, we implemented the normalized Hausdorff distance to ensure the geometric tolerance of the design. This metric is defined as the ratio between the measured Hausdorff distance between desired path A and the obtained path B to the diagonal of path A's bounding box:
    \begin{equation}
        NH(A,B)=\frac{H(A,B)}{Diag(A)}.
    \end{equation}
    where $H(A,B)$ is the Hausdorff distance and $Diag(A)$ represents the diagonal of path A's bounding box. This dimensionless metric can both intuitively and quantitatively evaluate the geometric tolerance of the generated curve, ensuring the feasibility of the resulting design.
    
\subsubsection{Dynamic Modeling}

    Assume that the angular velocity $\dot{q}$ and acceleration $\ddot{q}$ are known at any time $t\in[0, T]$. We can derive the linkages' velocity and acceleration using kinematic modeling. Therefore, the associated dynamic equations can be formulated as a set of second-order differential equations of the general form:
    \begin{equation}
        \tau =M(q)\ddot{q}+H(q,\dot{q}),
    \end{equation}
    where $\tau$ is the actuator torque, $M(q)$ is the mass matrix, and $H(q,\dot{q})$ are the forces that represent the centripetal, Coriolis, gravitational, and externally applied forces. Since we fabricate the link with lightweight 3D-printed parts and assemble the actuator proximal to the base, the dynamic effects caused by the mass matrix and acceleration term are negligible in this study. It is a reasonable simplification when the moving parts are light enough, and the generated motion is relatively slow. Additionally, ball bearings and lubrication are incorporated into the link joint design to minimize friction. Therefore, the simplified actuator torque $\tau$ can be determined by:
    \begin{equation}
        \tau(t) =J(q)^{T}F_e(t),
    \end{equation}
    where $F_e(t) \in \mathbb{R}^{6\times1}$ is the wrench applied on the linkage's end-effector $p$, and $J(q)$ is the Jacobian matrix in Eq.~\eqref{eq:jacobian}. This simplification is also helpful in ignoring the small weight change during the parameter optimization process. 
    
    Despite this simplification, the resulting actuator torque $\tau(t)$ provides a meaningful measure of the dynamic performance of the given linkage system under a specific workload. The role of dynamic modeling is to efficiently quantify the various linkage designs during the parameter optimization stage, rather than precisely simulating their dynamic behavior. In particular, our dynamic modeling aims to estimate the poor dynamic performance caused by the combination of design parameters and workload. Due to the singularity configuration, poor dynamic performance typically happens when the actuator cannot drive the end-effector effectively. We can avoid such a situation by calculating the required maximum and cumulative torque, thereby improving the energy efficiency of our design.

\subsubsection{Parametric Optimization}

    This section presents our approach to optimizing the kinematic design parameters of the overconstrained linkage that satisfy the high-level requirements mentioned in Section~\ref{sec:Problem}. We formulate the design problem as an optimization problem with finite decision variables and non-linear constraints. Taking the user-specified closed spatial curve $C_{A}(s), s \in[0,1]$ defined in the world coordinate as input, we aim to design the overconstrained linkage whose end-effector point $p$ can generate a spatial trajectory $C_{B}$ that can approximate the input curve. The design parameter $\mathcal{X}_{i}=[\phi _{i},\: p_{0},\: ^{i}T_{w}]$ includes link-related kinematic parameters $\phi _{i}$, end-effector point parameters $p_{0} \in \mathbb{R}^{3}$, and the homogeneous transform matrix from the world frame to the local one $^{i}T_{w} \in SE(3)$. Then, we can formulate the optimization problem as follows:
    \begin{equation}
       \begin{aligned}
        \min_{\mathcal{X}}   \quad & w_{1}E_{s}\left ( C_{A},C_{B} \right )+w_{2}\frac{\tau_{\max}}{\tau _{s}}+w_{3}\int_{0}^{T}\left | \tau (t)\omega (t) \right |dt\\
        \textrm{s.t.}   \quad    
        & \min_{t \in [0,T]} \left ( \kappa (J_{a}J_{a}^{T}) \right )\geq \xi  \\
          & p(T^{k})=o_{k}, \: \forall k\in [1,m] \\
          & p_{link}(t)\in R_{i}, \: \forall t \in [0,T]  \\
          & x_{L}\leq x \leq x_{U} \\
          &  NH(C_{A},C_{B})\leq \delta_{t}   \\
        \end{aligned}
        \label{eq:design_parameter_opt}
    \end{equation}
    where $E_{c}(C_{A}, C_{B})$ is the similarity metric (Eq. \ref{eq:similarity}) between the input and generated curves, $\tau_{max}$ is the maximum torque required to drive the overconstrained linkage, $\tau_{s}$ is the stall torque of the actuator. The last term in the objective function represents the total required mechanical energy consumption during the whole period $T$. Note that the first term in the objective function is related to the kinematic performance of the linkage mechanism, and the other two terms refer to its dynamic performance. $w_{1}$, $w_{2}$, and $w_{3}$ are set to 10, 1, and 0.1, respectively, to balance the impact of different terms for practical experiments. 
    
    While minimizing the given objective function, the design is also constrained by the listed formulation related to geometric tolerance, kinematics, workspace, and fabrication. First, the mechanism's minimal condition number $\kappa$ during the whole motion period $T$ should be no smaller than a threshold value, where $\xi=0.01$ in our experiments. Then, the end-effector point $p$ should follow the timing requirement specified by the key points $o_{k}$ and time interval $T^{k}$. Given a curve $C_{A, i}$ defined in the world coordinate system, the generated linkage design should be within the box boundary $R_{i}$ to avoid possible collisions between different mechanisms and adjust the integrated layout. Additionally, the kinematic design parameter $x$ should also be constrained by the lower boundary $x_{L}$ and upper boundary $x_{U}$, which are determined by both kinematic and fabrication considerations. Finally, the normalized Hausdorff distance $NH(C_{A}, C_{B})$ should no larger than a threshold value $ \delta_{t}$ (10$\%$ in this study), which ensures the geometric tolerance between the desired curve $C_{A}$ and the obtained one $C_{B}$.
    
    We uniformly sample 30 to 50 discrete points for practice to evaluate the objective function. Increasing the number of sampling points will slightly improve the objective function value, but at a much higher computational cost. We discretize all the constraints concerning time $t$ and resolve the optimization problem via a gradient-free method (i.e., the method of covariance matrix adaptation evolution strategy (CMA-ES)~\cite{hansen2003reducing}) by merging the constraints as the penalty terms to the objective function. The computing time of each design involved in this study is listed in Table~\ref{tab:Results-CompTime}. 
    \begin{table}[t]
        \centering
        \caption{Statistics of results presented in this study.}
        \label{tab:Results-CompTime}
        \begin{tabular}{ccccc}
        \hline
        \textbf{Figure} &
          \textbf{Name} &
          \textbf{\begin{tabular}[c]{@{}c@{}}Optim. \\ Dyn\end{tabular}} &
          \textbf{\begin{tabular}[c]{@{}c@{}}Para. Optim. \\ Time (min)\end{tabular}} &
          \textbf{\begin{tabular}[c]{@{}c@{}}Geom. Optim. \\ Time (min)\end{tabular}} \\ \hline
       \ref{fig:Results-KinExp}         & Heart                & No  & 9  & 8  \\ \hline
       \ref{fig:Results-DynExp} (left)  & Lifting Machine      & No  & 6  & 13 \\ \hline
       \ref{fig:Results-DynExp} (right) & Lifting Machine      & Yes & 13 & 15 \\ \hline
       \ref{fig:Results-BoatMan}        & Boat Man             & No  & 8  & 14 \\ \hline
       \ref{fig:Method-Overview}(c)     & \textit{OveRHex} (F) & Yes & 10 & 14 \\ \hline
       \ref{fig:Method-Overview}(c)     & \textit{OveRHex} (M) & Yes & 12 & 13 \\ \hline
       \ref{fig:Method-Overview}(c)     & \textit{OveRHex} (H) & Yes & 11 & 12 \\ \hline
       \ref{fig:Discussion-OtherDesign} & Bennett Design       & No  & /  & 8  \\ \hline
       \ref{fig:Discussion-OtherDesign} & Planar Design        & No  & /  & 3  \\ \hline
       \ref{fig:Discussion-OtherDesign} & Spherical Design     & No  & /  & 5  \\ \hline
       \ref{fig:Discussion-OtherDesign} & Goldberg Design      & No  & /  & 18 \\ \hline
        \end{tabular}
    \end{table}

\section{Results}
\label{sec:Result}
\subsection{Evaluation of Kinematic Performance}
\label{sec:Result-Kine}

\begin{figure}[ht]
        \centering
        \includegraphics[width=1\linewidth]{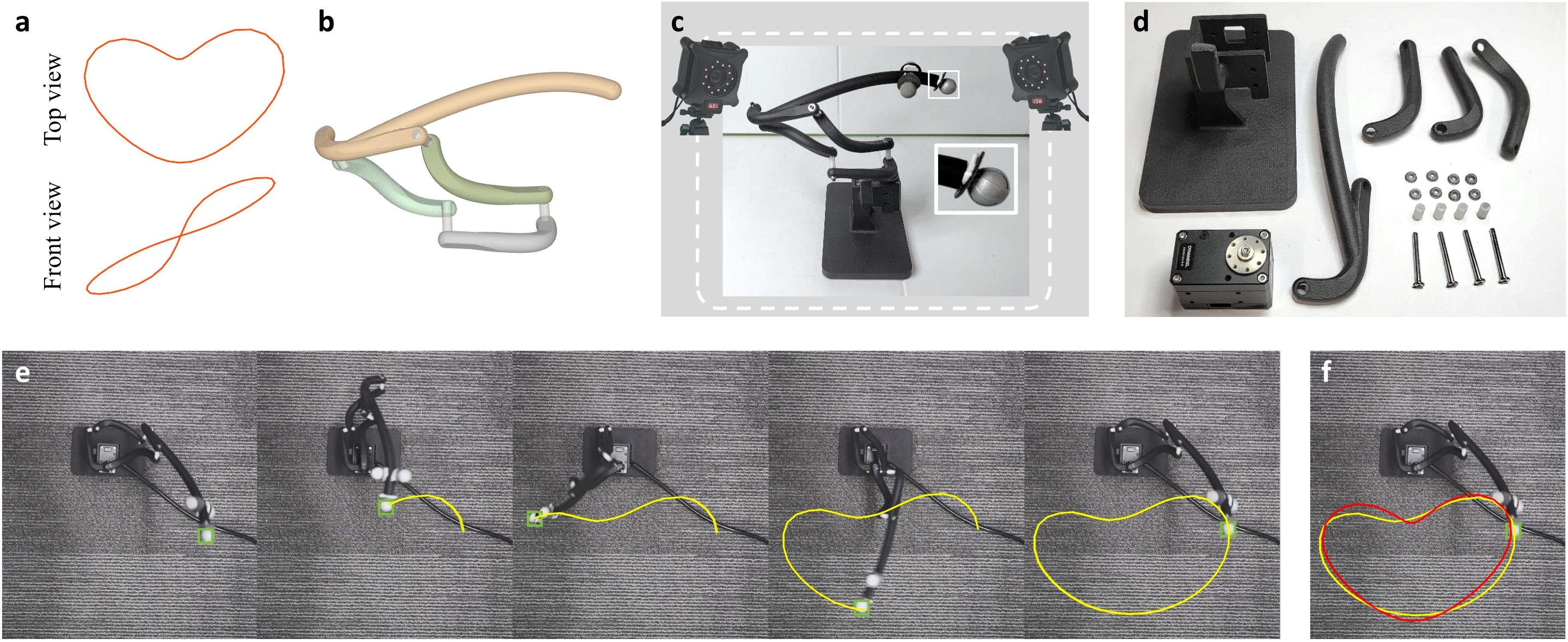}
        \caption{Evaluating the kinematic performance of our overconstrained mechanism for realizing a spatial HEART curve. (a) The target curve in top and front views, respectively; (b) our designed overconstrained linkage mechanism; (c) and (d) 3D printed overconstrained limb and its parts, where a marker is attached at the end-effector; (e) tracking the marker via motion capture cameras; and (f) comparing the captured curve (yellow) and the target curve (red) projected onto the image.}
        \label{fig:Results-KinExp}
    \end{figure}

    We validate our computational design framework in hardware to evaluate the kinematic performance of our linkage system. This experiment takes a complicated spatial motion as input, whose projection in the $x$-$y$ plane is a heart-like shape, while in the $y$-$z$ plane, it is an infinity symbol; see Fig.~\ref{fig:Results-KinExp}(a). 
    Our computational design framework generates a collision-free, overconstrained linkage mechanism that realizes the target curve using a single actuator. The resulting design is fabricated using the 3D printing technique, where the links and support parts are printed separately and assembled to form the final mechanism. The revolute joint consists of a steel shaft and flanged ball bearings, ensuring assembly accuracy and smooth motion. All the revolute joints have the exact specification and thickness, except the active joints, which need additional structure to connect with the actuator's output shaft.

    In the hardware experiment, we use the motion capture technique to record the end-effector's trajectory via an attached marker. In this regard, 12 motion-capture cameras\footnote{\url{https://en.nokov.com/products/motion-capture-cameras/Mars.html}} track and record the position of the marker at 380 FPS. As illustrated in Fig.~\ref{fig:Results-KinExp}(c), the tracking marker is attached to the follower end-effector of our design. Figure~\ref{fig:Results-KinExp}(e) visualizes the motion of the marker for a whole motion period. The green box highlights the marker, and the solid yellow line is the motion generated by the mechanism. To quantitatively compare the tracked and target curves, we transfer the 3D target curves from the world frame to the image frame via the camera projection matrix. Figure~\ref{fig:Results-KinExp}(f) shows the comparison result, where the modified spatial curve similarity metric between the two curves is 0.045. Additionally, given the bounding box of the spatial curve with a diagonal of 321.6 mm, the normalized Hausdorff distance is $22.5/321.6=6.9\%$, indicating that the resulting design is physically feasible. The inaccuracy of assembly, fabrication, and object tracking might cause this difference.

\subsection{Evaluation of Dynamic Performance}
\label{sec:Result-Dyna}

    To validate the dynamic performance of our design, we conducted a set of experiments comparing an optimized overconstrained linkage design with a baseline design that only considers the similarity metric as the objective function. We adopt the same target curve, workload, and initial setting as input to ensure consistency. More specifically, the input curve is a warped ellipse whose load condition is to resist a constant downward force of 5N applied on the end-effector. The only difference between the two experiments is the weight terms $w_2$ and $w_3$ of the objective function in Eq.~\eqref{eq:design_parameter_opt}, which are set to zeros for the baseline case. Therefore, the baseline design only considers the similarity between the generated motion and the target one, while the optimized design considers both the curve similarity and dynamic performance.

    We found that both resulting designs can approximately achieve the target motion with the given workload. While the baseline design had a relatively small similarity metric (0.0135) compared with the optimized one (0.0149), indicating a slight kinematic performance enhancement (10$\%$). In contrast, the theoretical calculation results showed that the cumulative mechanical energy cost of the baseline design was approximately 1.39 \unit{\joule} per cycle, which was about 30\% larger than that of the optimized design (1.01 \unit{\joule} per cycle). To verify the theoretical analysis, we conducted simulation experiments with different linkages and identical task requirements. The simulation is implemented in MATLAB within the Simulink environment on Windows, and the computations are performed on a single core of a 3.20 GHz Intel i9 processor. During the experiments, we applied an external force to the end-effector of the linkages and recorded the required actuation torque. 

    \begin{figure}[t]
        \centering
        \includegraphics[width=\columnwidth]{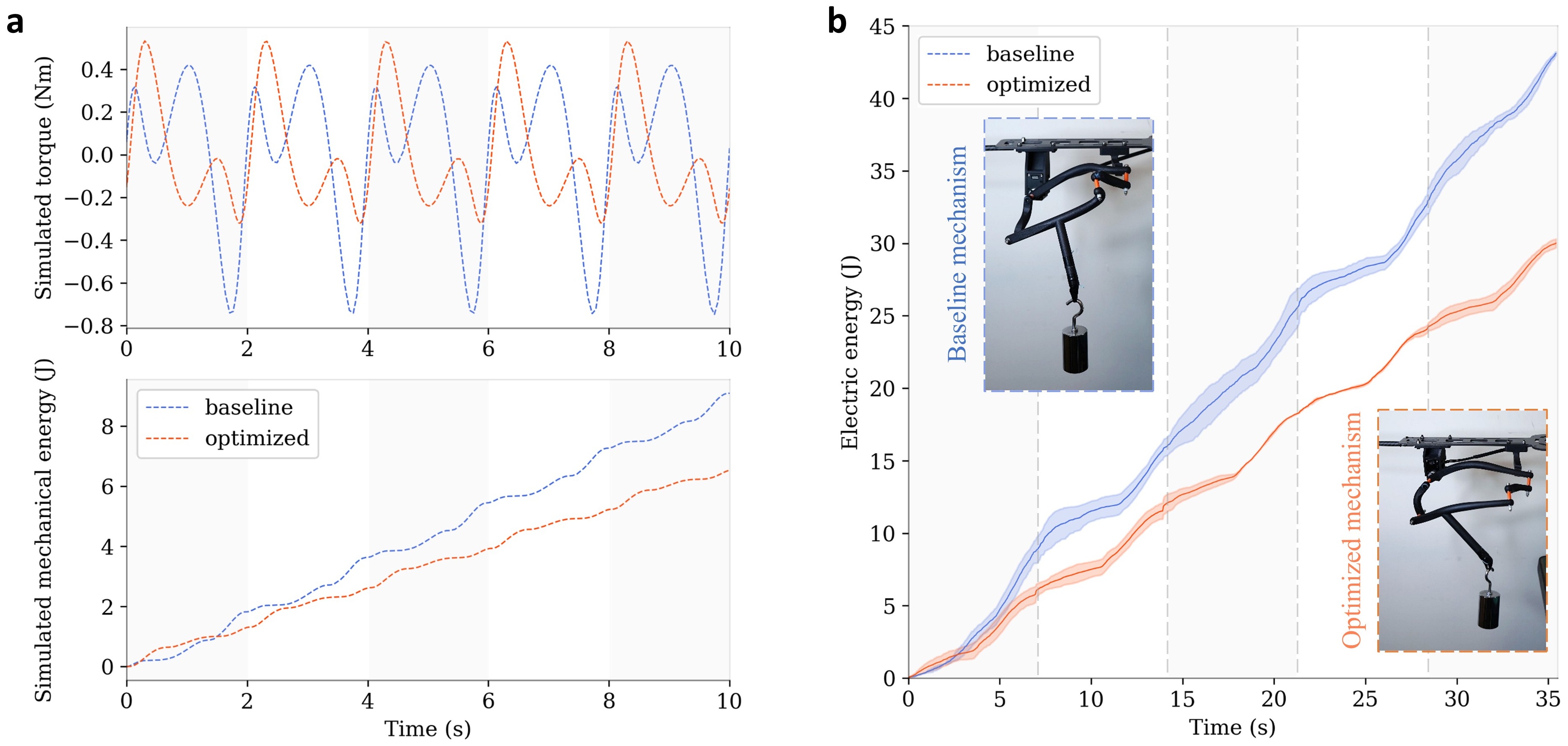}
        \caption{Experiment to validate dynamic performance. (a) Simulation experiment result. We compared the required actuation torque and mechanical energy cost between the optimized design (red line) and the baseline one (blue line) in the simulation environment. (b) Hardware experiment. (Bottom right) An overconstrained limb optimized by our approach. (Top left) An overconstrained limb generated by a baseline approach only concerns the similarity metric. Compare the energy consumption of two limbs for the same target curve and payload (500 \unit{\gram}).}
        \label{fig:Results-DynExp}
    \end{figure}
    
    As illustrated in Fig.~\ref{fig:Results-DynExp}(a), the simulated maximum torque of the baseline design was 0.74 \unit{\newton\meter} during the whole period, which is 27$\%$ larger than the optimized one (0.53 \unit{\newton\meter}) when driving the same workload. Additionally, the simulated mechanical energy cost was reduced from 9.10 \unit{\joule} (baseline) to 6.53 \unit{\joule} (optimized), indicating a significant improvement in energy efficiency (28$\%$) after considering the dynamic-related metric. Therefore, our optimized mechanism performed well when considering the kinematic and dynamic behavior comprehensively. We found that our method primarily achieves efficiency by reducing the maximum torque required for actuation, a finding also observed by  Megaro et al.~\cite{megaro2017computational}. Here, the dynamic-related metric enforced design parameters away from the singularity posture, even though it may reduce the curve similarity to a certain degree. At the same time, designers might prefer sacrificing part of the curve's similarity to significantly improve its dynamic performance. Additionally, designers can adaptively adjust the weight terms of the objective function based on specific application scenarios and workload.

    To verify the simulation results, we present hardware experiments that compare the actual energy costs of the baseline and optimized designs for completing the same task. Here, we fabricate the above linkage mechanism design via 3D printing and assemble it. All the links are printed with the same materials, including the rod radius and the size of the revolute joint. Additionally, we maintain each design, which is driven by a single Dynamixel XM-430 servo and supplied with the same stabilized voltage power source. As shown in Fig.~\ref{fig:Results-DynExp}(b), we fix the linkage mechanism on a plate and hang a 500 \unit{\gram} weight on the follower's end-effector as the workload. 

    We enforce the end-effector to move uniformly and slowly via inverse kinematics to minimize dynamic effects. Our prototype received the control command from a tethered computer during the experiment and returned the current feedback. Since the actuator does not have a torque sensor, the measured mechanical energy cost can be equivalently represented by the cumulative electric energy consumption, calculated by multiplying the current feedback by the input voltage and the discrete time interval. Figure~\ref{fig:Results-DynExp} presents the energy consumption for the prototype to replay the same task. For each task, we reproduce the entire motion period five times, where the solid line represents the average energy cost value. We find that the results of real-world physical experiments are consistent with those of the simulations, exhibiting the same trend. Compared to the energy consumption of the baseline design, the optimized design has a 32$\%$ decrease, which is similar to the forecasted one. Therefore, the hardware experiments indicate that the dynamic metric term plays a crucial role in the performance of practical designs. On the other hand, it demonstrates that our mechanism configuration can realize spatial motion with improved dynamic performance for large workloads, indicating its potential for further engineering applications.

\subsection{Boat Man}
\label{sec:Result-BoatMan}

\begin{figure}[t]
        \centering
        \includegraphics[width=1\linewidth]{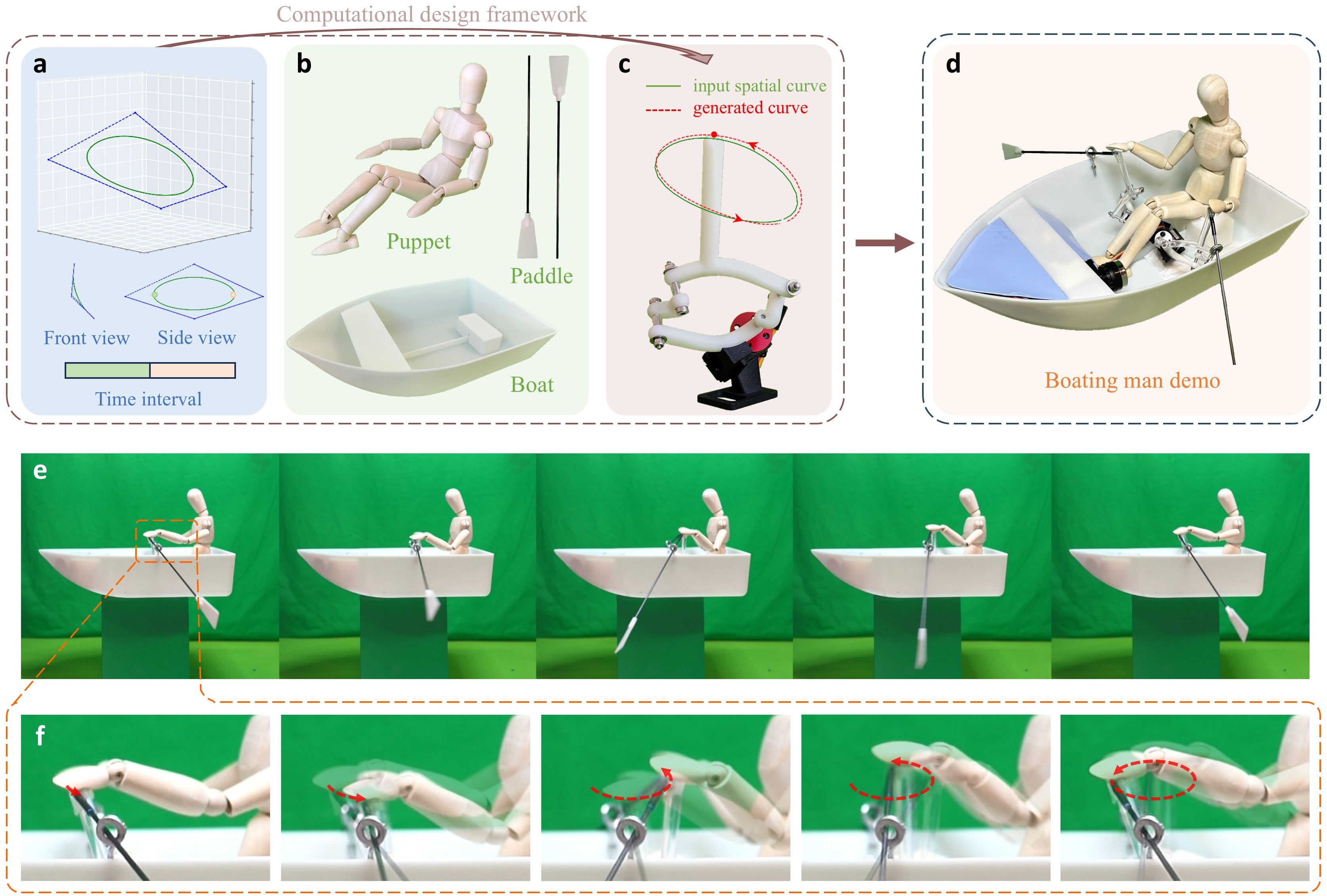}
        \caption{Mechanical toy designed by our approach. (a) The motion trajectories, (b) other parts, (c) our overconstrained mechanism, (d) the final design, (e) the image sequence of the mechanical toy, and (f) the motion of the end-effector.}
        \label{fig:Results-BoatMan}
    \end{figure}

    Our computational design framework can be used to design mechanical automata with specific motion behavior by generating overconstrained mechanisms (see Fig.~\ref{fig:Results-BoatMan}). Thanks to the precise spatial-temporal motion control provided by our overconstrained linkages, we can design the mechanical toy with a realistic and robust performance by adjusting the motions of different end-effectors. In Fig.~\ref{fig:Results-BoatMan}, we design and fabricate a mechanical toy that mimics a man who is rowing his boat, called \textit{boat man}. We define the cyclic spatial trajectory in Fig.~\ref{fig:Results-BoatMan}(a) as the input motion to mimic the boating scenario; then, our technique can generate the corresponding collision-free linkage mechanism to approximately track the input motion via its end-effector. As shown in Fig.~\ref{fig:Results-BoatMan}(c), the green solid line represents the desired spatial curve given by the user, and the red dashed line represents the generated one by the end-effector of our linkage mechanism. Then, we can obtain the mechanical toy \textit{boat man} by assembling other parts listed in Fig.~\ref{fig:Results-BoatMan}(b). Figure~\ref{fig:Results-BoatMan}(e) presents five boating poses as part of the full-cycle motion, and Fig.~\ref{fig:Results-BoatMan}(f) zooms in on the motion of the hand, indicating the precise motion tracking of our design. Note that the generated motion is a spatial curve whose side view resembles an ellipse, better imitating the actual boating function.

    Additionally, we fabricate the end-effector link using a transparent resin material via a 3D printing technique to enhance the visual performance of the final design. Furthermore, we implement temporal control to obtain more realistic behavior. Similarly, designers can take almost any closed cyclic motion with finite control points as input to design interesting mechanical toys with various behaviors, such as boxing, ping-pong, and skiing. Note that when multiple target curves exist, the designer should properly divide the free space into several boundary boxes for each linkage system at the beginning of the design stage to obtain collision-free motion.

\subsection{OveRHex Walker}
\label{sec:Result-OveRHex}

    We leverage our approach to design a hexapod robot that walks following a bio-inspired gait to demonstrate the engineering potential of our mechanisms. The gait is defined by six spatial curves representing the stick insect's trajectories of the legs~\cite{cruse2009principles}. According to the investigations of stick insects~\cite{cruse1995movement}, the front, middle, and hind movements are diverse spatial trajectories. In detail, the stance phase trajectory of the front and hind legs is inward, while one of the middle legs is outward. All swing phase trajectories are outward during the step cycle (see Fig.~\ref{fig:Method-Overview}). Note that the walking speed will also impact the shape of the gait, and here, we choose the slow walking gait as the input. Also, the time requirement is implemented according to the gait (e.g., moving slowly during the stance phase and quickly during the swing phase). In this regard, the leg trajectories on one side have been assigned three key points with the same time intervals, $\Delta T=T/3$, while the three legs on the other side have a symmetrical arrangement. To ensure the equilibrium of the hexapod robot during locomotion, we implement the tetrapod gait~\cite{cruse2006control} such that there are always at least four legs touching the ground. Taking the specified curve as input, our computational design framework generates three sets of linkage-based limbs, symmetrically arranged, as the legs of the bio-inspired hexapod robot. Each robotic leg features a well-designed, collision-free, and invertible linkage mechanism, actuated by a single servo motor. We also consider the work conditions caused by the ground reaction force in energy-efficient locomotion. As shown in Fig.~\ref{fig:Method-Overview}(a), we apply a constant vertical force on the gait trajectories when the leg touches the ground as the externally applied wrench.

    \begin{figure}[t]
        \centering
        \includegraphics[width=1\columnwidth]{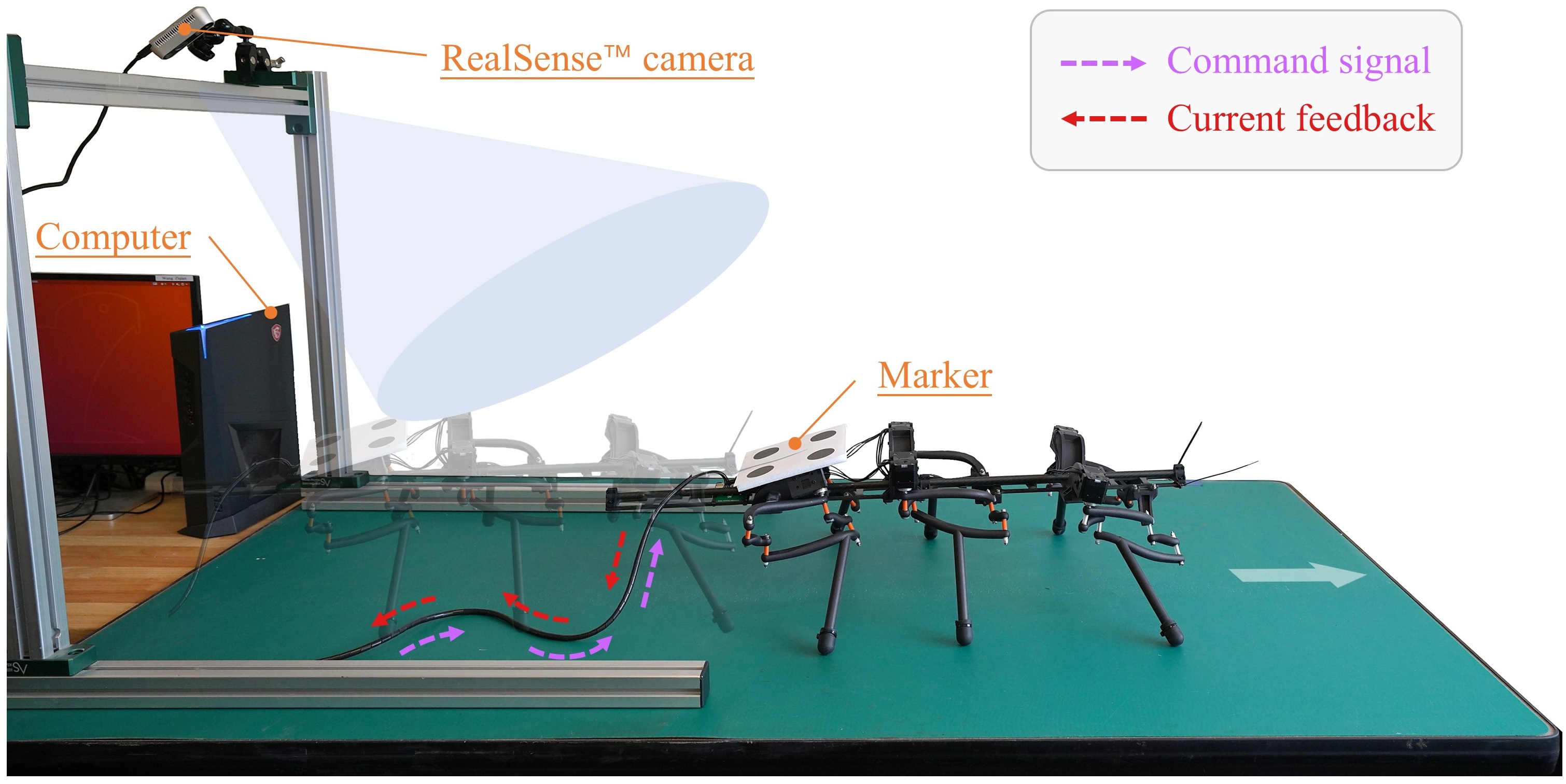}
        \caption{Experiment setup for the OverRHex locomotion.}
        \label{fig:Results-ExpSetup}
    \end{figure}

    As mentioned in Eq.~\eqref{eq:design_parameter_opt}, our computational design framework considers both the kinematic similarity of the input and its dynamic performance. Therefore, in addition to satisfying the curve similarity between input and output, our design will reduce the expected maximum torque and accumulated mechanical energy consumption, allowing the resulting robot to be driven efficiently by a micromotor. To verify whether our designed hexapod robot can walk, we fabricate and assemble a prototype hexapod robot using a 3D printing-based approach, as shown in Fig.~\ref{fig:Results-ExpSetup}. The mechanism link and actuator base are manufactured from Nylon 12 for their mechanical strength. One Dynamixel XM-430 W-210-R servo drives each leg with 3.0 \unit{\newton\meter} stall torque at a 12.0 V power supply. We used lubricated ball bearings to support the rotating joints, ensuring smooth movement. Furthermore, the leg tips (end-effectors) are coated with a silicone case (Dragon Skin 10) to prevent slipping. 

\begin{figure}[ht]
        \centering
        \includegraphics[width=1\linewidth]{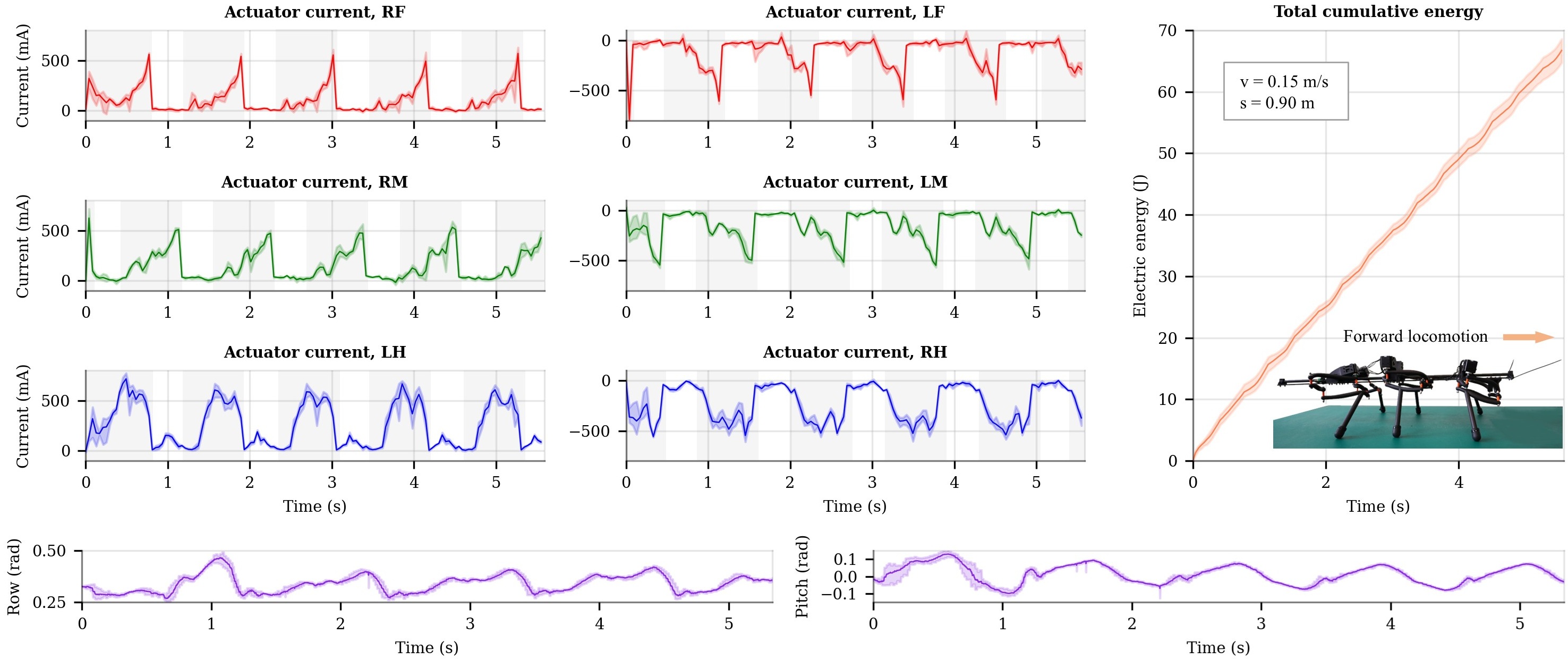}
        \caption{Hardware experiment results include the current value of each limb, roll, pitch angle, and electric consumption during locomotion experiments.}
        \label{fig:Results-LocoData}
    \end{figure}

\begin{figure}[ht]
        \centering
        \includegraphics[width=1\linewidth]{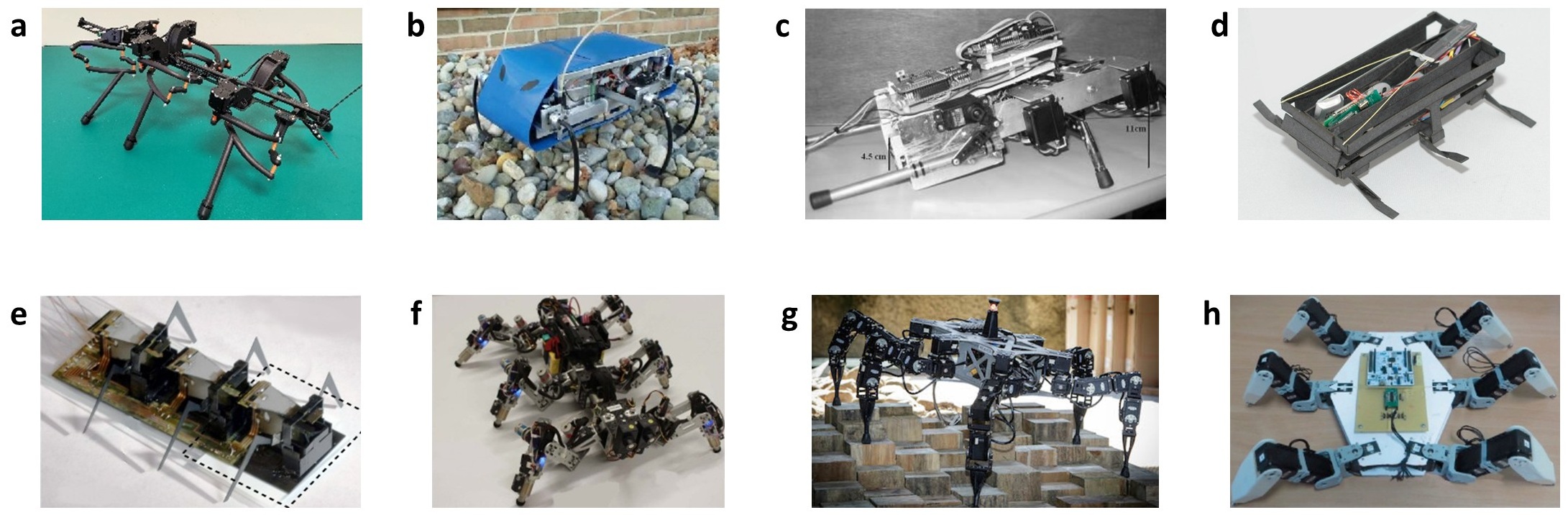}
        \caption{Hexapod robots in recent years. (a) OveRHex; (b) RHex; (c) GREGOR \uppercase\expandafter{\romannumeral1}; (d) DASH; (e) HAMR; (f) AMOS; (g) Weaver; (h) HexaV4.}
        \label{fig:Results-HexapodRobots}
    \end{figure}

    To validate the engineering potential of our design, we conducted a locomotion experiment to measure energy consumption and stability during the forward walking task in Fig.~\ref{fig:Results-ExpSetup}. The tethered computer could send the control signal and receive and record the servos' angular and current feedback during the task. Meanwhile, the deep camera (Intel RealSense) above would also record the position and pose of the hexapod robot via the attached marker~\cite{garrido2014automatic}. To ensure the consistency of experiment results, we performed the same trial five times with the identical setup. As shown in Fig.~\ref{fig:Results-LocoData}, we present the hardware experiment results, with the solid lines representing the mean value and the light-colored region representing the variance of repeated trials, including the current feedback and the row and pitch angle of the body. 
    Regarding the hardware experiment, the total cumulative energy is calculated by accumulating the product of the servos' instantaneous electric power and the discrete-time interval. Typically, the energy efficiency of a walking robot is measured by \textit{Cost of Transport} (COT), which is also known as specific resistance~\cite{gabrielli1950price}, and COT is defined as follows:
    \begin{equation}
        COT=\frac{E}{mgd}=\frac{P}{mgv},
        \label{eq:COT}
    \end{equation}
    where $E$ is consumed energy [\unit{\joule}], $m$ is the mass of the robot [\unit{\kilogram}] (i.e., 2.3 \unit{\kilogram} for OveRHex), $g$ is gravitational acceleration [\unit{m/s^2}], $d$ is the locomotion distance [\unit{\meter}]. While $P$ is consumed power [\unit{\watt}] and $v$ is walking velocity [\unit{m/s}]. A smaller COT indicates higher energy efficiency. For each successful locomotion, the energy consumption $E$ is calculated based on the electrical current feedback from all motors of our design, and the walking distance $d$ is 0.9 \unit{\meter} with time $t=5.2$ \unit{\second}. Therefore, by using the measured power and the distance, the COT of each experiment was calculated. Table~\ref{tab:COT} shows that the COT of the OveRHex robot is about 3.25 at walking velocity $v=0.15$ \unit{m/s}. 
    \begin{table}[ht]
        \centering
        \caption{Cost-of-Transport (COT) of small-sized hexapod legged robots.}
        \label{tab:COT}
        \begin{tabular}{cccccc}
        \hline
        \textbf{\begin{tabular}[c]{@{}c@{}}Hexapod\\ Robot\end{tabular}} &
          \textbf{COT} &
          \textbf{\begin{tabular}[c]{@{}c@{}}Velocity\\ (\unit{m/s})\end{tabular}} &
          \textbf{\begin{tabular}[c]{@{}c@{}}Mass\\ (\unit{\kilogram})\end{tabular}} &
          \textbf{DoF} &
          \textbf{\begin{tabular}[c]{@{}c@{}}Actuator\\ Type\end{tabular}} \\ \hline
        \begin{tabular}[c]{@{}c@{}}\textbf{OveRHex}\\\textbf{(This Work)}\end{tabular}       & \textbf{3.25} & 0.15 & 2.30   & 6  & Servo \\ \hline
        \begin{tabular}[c]{@{}c@{}}RHex\\\cite{saranli2001rhex}\end{tabular}         & 3.74-14       & 0.55 & 7.00   & 6  & DC    \\ \hline
        \begin{tabular}[c]{@{}c@{}}GREGOR1\\\cite{arena2006realization}\end{tabular} & 70            & 0.03 & 1.20   & 16 & Servo \\ \hline
        \begin{tabular}[c]{@{}c@{}}DASH \\\cite{birkmeyer2009dash}\end{tabular}      & 14.7          & 1.50 & 0.0162 & 1  & DC    \\ \hline
        \begin{tabular}[c]{@{}c@{}}HAMR\\\cite{baisch2010biologically}\end{tabular} &
          128 &
          0.23 &
          0.002 &
          6 &
          \begin{tabular}[c]{@{}c@{}}Piezo-\\electric\end{tabular} \\ \hline
        \begin{tabular}[c]{@{}c@{}}AMOS\\\cite{xiong2015adaptive}\end{tabular}       & 3.4-11.7      & 0.12 & 5.40   & 19 & Servo \\ \hline
        \begin{tabular}[c]{@{}c@{}}Weaver\\\cite{bjelonic2018weaver}\end{tabular}    & 10            & 0.35 & 7.03   & 30 & Servo \\ \hline
        \begin{tabular}[c]{@{}c@{}}HexaV4\\\cite{luneckas2021hexapod}\end{tabular}   & 5.75          & 0.33 & 1.50    & 18 & Servo \\ \hline
        \end{tabular}
    \end{table}
    Moreover, our technique enables the resulting design to achieve more energy-efficient locomotion (see Fig.~\ref{fig:Results-HexapodRobots} and Table~\ref{tab:COT}), compared to other small-sized hexapod robots. Note that the lowest COT value does not mean that our robot is comprehensively superior to others; rather, it indicates that the resulting design has significant engineering potential and outstanding performance criteria in terms of energy efficiency to a certain degree. In this experiment, our method primarily achieves energy efficiency by reducing the number of actuation (one degree of freedom per limb) and the number of mechanical elements. On the other hand, the previous work has also demonstrated the potential advantage of overconstrained robotic limb design in promoting energy efficiency for locomotion tasks~\cite{gu2023computational}. We believe that the proposed framework enables designers to develop energy-efficient engineering designs with overconstrained linkages and formulate novel robot structures that utilize fewer actuators while enhancing dexterity.

\section{Discussion}
\label{sec:Discuss}

\subsection{Generalized to Linkage-Based Mechanisms}
\label{sec:Discuss-LinkGen}

\begin{figure}[t]
        \centering
        \includegraphics[width=0.8\columnwidth]{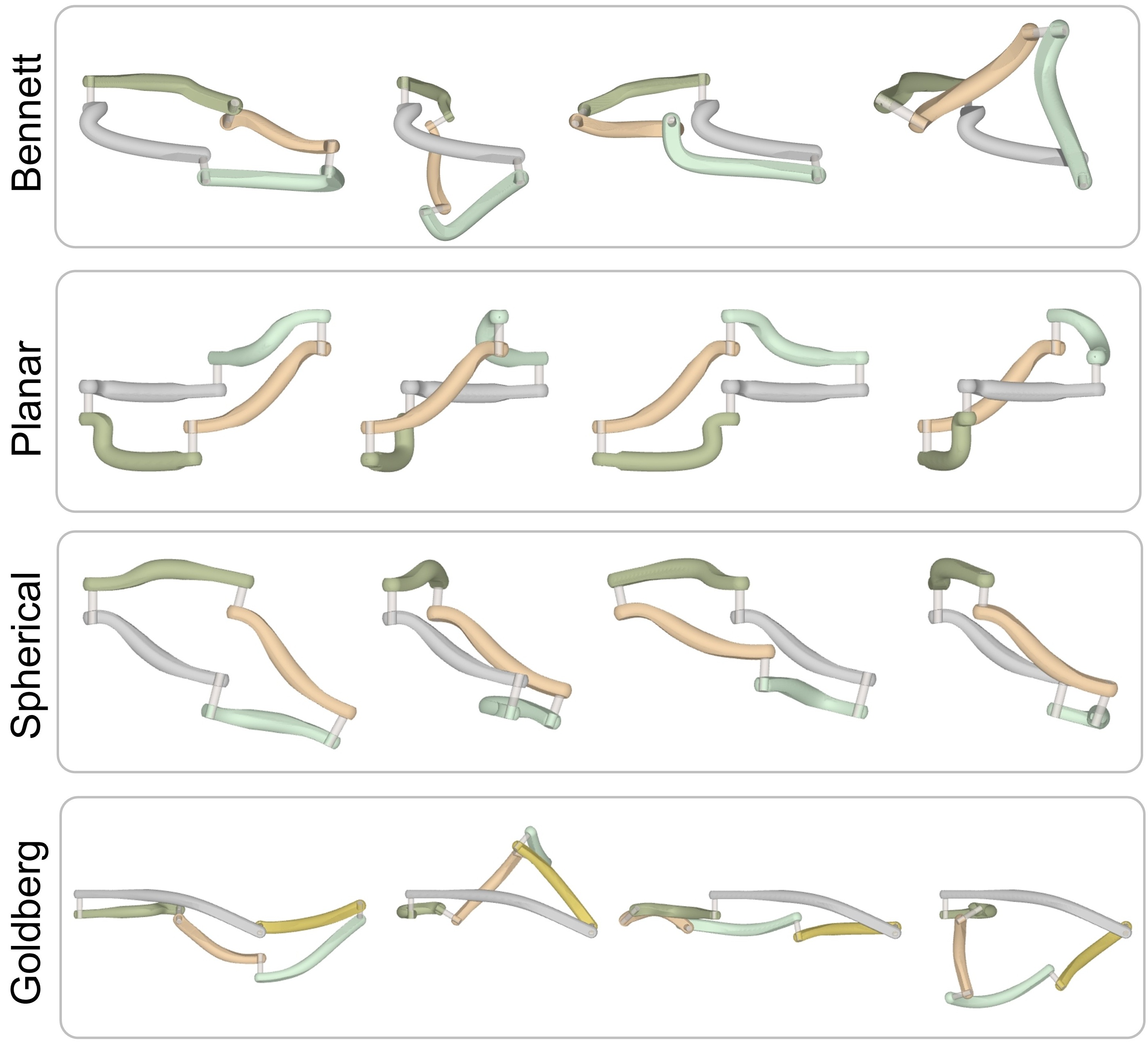}
        \caption{Four linkage mechanisms in different topology configurations designed by our approach to avoid self-collision.}
        \label{fig:Discussion-OtherDesign}
    \end{figure}

    The proposed self-collision-avoidance method in this study can be generalized to nearly any linkage-based mechanism design process in robotics applications. Once the forward kinematics is provided, our method can formulate the involved link elements into a parametric deformable form, calculate the collision energy term, and rapidly optimize the geometric shape via the gradient information. Therefore, explicit kinematic derivation is essential to this study. One possible solution is to utilize the equivalent open-chain technique, which involves deriving the explicit representation of the input joint angle in terms of other joint angles from the closure equation of the mechanism and formulating the forward kinematics using the product of exponentials (POE) formula. Compared to resolving the linkage kinematics as a sub-optimization problem by minimizing the non-linear least squares of the constraints, the explicit formulation can significantly reduce computing time in subsequent stages. To demonstrate the above discussion, we implement our method to design various linkage-based mechanisms, including the planar, spherical, and Goldberg linkage (5R case). Note that here, we select planar and spherical linkages with equivalent link length parameters as the optimization objective to increase the difficulty of the design process, since our algorithm should not only consider the collision among linkages but also avoid the joints colliding with each other. As illustrated in Fig.~\ref{fig:Discussion-OtherDesign}, our computational design framework generates collision-free designs in a staggered manner. 
    Compared to standard designs with straight linkages, the curved ones result in more compact designs that occupy less space during full-cycle motion. On the other hand, our method can be extended to design complex linkage systems with more link elements. Here, the geometric shape of a 5R Goldberg linkage was optimized to complete collision-free full-cycle motion. The increased number of linkages may generate more complicated trajectories of the end-effector, while a longer computing time would be required. To sum up, based on the extension of implementing the technique to other linkage mechanisms with different configurations and link numbers, we demonstrate that our design method is consistent with a typical design process, and can be easily generalized to handle the self-collision-avoidance design problem of various linkage-based mechanisms, providing a novel solution for non-expert designers.

\subsection{Towards Energy-Efficient Robotic Limb Design}
\label{sec:Discuss-LimbDes}

    Energy efficiency has been a key factor in the development of robotic limb design, particularly in terms of the electrical power of different motors and control units, the mechanical power of actuating body mechanisms, and numerous other factors. To ensure the energy efficiency of the resulting robotic limb design, the proposed computational framework in this study mainly leverages the following efforts: 1) reducing the actuation; 2) utilizing overconstrained linkages; 3) optimizing energy-related metrics; 4) reproducing bio-inspired spatial trajectories. 
    
    First, we leverage the design complexity to reduce actuation. Specifically, although a robotic arm can achieve most tasks with a relatively simple structure and multiple actuators, our design can perform approximate repetitive motion via the overconstrained linkage element driven by a single actuator. This significantly reduces the weight of actuators, parts, and moving components, resulting in lower energy consumption. Furthermore, the specific design can reduce the complexity of the control system, not only being easy to control due to fewer actuators, but also eliminating the need to implement a self-collision detection algorithm, making it potentially widely applicable in many applications. 
    
    Secondly, in this study, we utilize the overconstrained linkage as the driving element due to its relatively simple topology structure (a revolute-only spatial closed-loop motion chain) and its outstanding mechanical properties. In detail, compared with the widely used serial and planar configuration for mechanical transmission, the overconstrained spatial linkage might be able to improve the energy efficiency for specific tasks by introducing the additional design parameters (twist angle of links), which have been demonstrated in a previous study~\cite{gu2023computational}. In this regard, the overconstrained robotic limbs achieved enhanced lateral locomotion and turning performance compared to their serial and planar counterparts. This suggests the potential advantage of enhancing energy efficiency for omnidirectional locomotion. 
    
    Thirdly, we consider the energy-related metric during the parameter optimization stage. As illustrated in Fig.~\ref{fig:Results-DynExp}, by adding the dynamic term to the objective function, we demonstrate that the optimized design significantly reduces energy consumption compared to the baseline (30$\%$) for completing a similar task. Last, compared with simplifying to a 2D curve and using a planar mechanism, our design framework can directly approximate the bio-inspired spatial trajectories and reproduce them straightforwardly. While the locomotion experiment properly demonstrates this. Besides the factors mentioned above, the bio-inspired trajectories and gaits play a crucial role in creating a prototype hexapod robot with outstanding energy efficiency in forward locomotion. With our computational framework, designers may be able to reproduce bio-inspired behaviors, such as locomotion, swimming, and flapping, which have evolved to be energy-efficient over time. 
    
    The proposed framework enables designers to generate feasible and energy-efficient limb designs tailored to specific tasks. It provides instructive insights into determining the geometric shape of advanced robotic limb designs. We acknowledge the limitations of the energy-related conclusion in this prototype study, as numerous factors (such as mechanical design and control algorithm) that may impact a robot system's energy efficiency are outside the scope of this study and will be addressed in future research.

\section{Final Remarks}
\label{sec:Final}

    This study proposes a computational design framework for generating 1-DoF overconstrained robotic limbs with energy-efficient and self-collision-free motion. Through the parameterized representation of link elements and a continuous collision energy function, we can optimize the geometric shape of almost any linkage-based mechanism to achieve self-collision-free motion. Taking the Bennett linkage as an example, we formulate a spatial mechanism synthesis problem with an overconstrained linkage. Given the spatial motion trajectories and workload as input, our computational design framework can generate an overconstrained robotic limb to approximate the input motion driven by a single actuator. Our algorithm considers both spatial curve similarity and energy-related terms to ensure the resulting design completes the given task energy-efficiently. We fabricate and assemble the overconstrained robotic limb design for validation and implement several hardware experiments. Kinematic performance is evaluated by realizing a spatial heart-like trajectory. Compared with the input spatial trajectory, the measured data of our designed overconstrained limb shows a relatively small normalized Hausdorff distance value of $6.9\%$. The dynamic evaluation experiment demonstrates that the optimized robotic limb reduces the energy cost by approximately 30$\%$ compared to the baseline design for the exact task requirement.

    Additionally, our technique can be used to design mechanical characters, such as the \textit{boat man}. Finally, we use the bio-inspired gait trajectories as input, optimize, and fabricate the prototype hexapod robot \textit{OveRHex} with only one actuator per leg. Thanks to the reduction of actuation, overconstrained limb, energy-related optimization, and spatial gait trajectories inspired by nature, the proposed robot achieves outstanding energy efficiency in forward locomotion, with a State-Of-The-Art COT value (3.25) among small-size hexapod robots in recent years. We believe that our framework can help designers generate physically feasible and energy-efficient overconstrained limb designs for specific task requirements in advanced robotic applications, or provide instructive insights into determining the geometric parameters of robotic limbs to avoid self-collision.

    This study has several limitations that highlight interesting directions for future work. First, our overconstrained linkage can only perform smooth and relatively simple motions due to the finite design parameters and topology configuration. For cases where the design candidates cannot satisfy the geometric tolerance constraint, one possible solution is to combine our overconstrained mechanism with other mechanisms, such as non-circular gears~\cite{xu2020computational}. The proposed formulation can still handle these cases. Additionally, we acknowledge that optimizing a central axis design by minimizing its total link length or link curvature may be a promising research direction. Instead of the polynomials, utilizing an advanced curve representation method (such as a B-spline curve) would be a preferred choice when implementing further geometric optimization techniques to the link design, which will be addressed in future work. The uniform cross-section bar is not the most optimal structure in terms of bounding volume and mechanical performance, such as stiffness. Therefore, one possible direction for future work is to incorporate additional geometric optimization techniques, such as topology optimization~\cite{shin2023topology}, to enhance the mechanical characteristics after determining the collision-free design volume of each link element. Furthermore, our overconstrained limb has various potential applications in advanced robotics, such as the flipping mechanism~\cite{abas2016flapping} and eyeball control mechanism~\cite{khan2018design}.

\section*{Acknowledgment}
\label{sec:Acknowledgment}

    This work was supported by the National Natural Science Foundation of China [grant number 62473189].

\section*{Supplementary Materials}
\label{sec:SupMat}

    \begin{itemize}
        \item \textbf{Movie S1}. \href{https://bionicdl.ancorasir.com/?p=1668}{Evaluation of Kinematic Performance}.
        \item \textbf{Movie S2}. \href{https://bionicdl.ancorasir.com/?p=1668}{Evaluation of Dynamic Performance}.
        \item \textbf{Movie S3}. \href{https://bionicdl.ancorasir.com/?p=1668}{Boat Man}.
        \item \textbf{Movie S4}. \href{https://bionicdl.ancorasir.com/?p=1668}{OverRHex Walker}.
    \end{itemize}

\bibliographystyle{unsrt}
\bibliography{References}  

\begin{thebibliography}{10}

\bibitem{biswal2021development}
Priyaranjan Biswal and Prases~K Mohanty.
\newblock \href{https://doi.org/10.1016/j.asej.2020.11.005}{Development of
  quadruped walking robots: A review}.
\newblock {\em Ain Shams Eng. J.}, 12(2):2017--2031, 2021.

\bibitem{kebria2016kinematic}
Parham~M Kebria, Saba Al-Wais, Hamid Abdi, and et~al.
\newblock \href{https://doi.org/10.1109/SMC.2016.7844896}{Kinematic and dynamic
  modelling of UR5 manipulator}.
\newblock {\em In: Proc. IEEE SMC}, pages 4229--4234, 2016.

\bibitem{ollero2021past}
Anibal Ollero, Marco Tognon, Alejandro Suarez, and et~al.
\newblock \href{https://doi.org/10.1109/TRO.2021.3084395}{Past, present, and
  future of aerial robotic manipulators}.
\newblock {\em IEEE Trans. Robot.}, 38(1):626--645, 2021.

\bibitem{riboli2023collision}
Marco Riboli, Matthieu Jaccard, Marco Silvestri, and et~al.
\newblock \href{https://doi.org/10.1016/j.robot.2023.104534}{Collision-free and
  smooth motion planning of dual-arm cartesian robot based on B-spline
  representation}.
\newblock {\em Robot. Auton. Syst.}, 170:104534, 2023.

\bibitem{hutter2016anymal}
Marco Hutter, Christian Gehring, Dominic Jud, and et~al.
\newblock \href{https://doi.org/10.1109/IROS.2016.7758092}{Anymal - A highly
  mobile and dynamic quadrupedal robot}.
\newblock {\em In: Proc. IEEE/RSJ IROS}, pages 38--44, 2016.

\bibitem{hoeller2024anymal}
David Hoeller, Nikita Rudin, Dhionis Sako, and et~al.
\newblock \href{https://doi.org/10.1126/scirobotics.adi7566}{Anymal parkour:
  Learning agile navigation for quadrupedal robots}.
\newblock {\em Sci. Robot.}, 9(88):eadi7566, 2024.

\bibitem{kuindersma2016optimization}
Scott Kuindersma, Robin Deits, Maurice Fallon, and et~al.
\newblock \href{https://doi.org/10.1007/s10514-015-9479-3}{Optimization-based
  locomotion planning, estimation, and control design for the atlas humanoid
  robot}.
\newblock {\em Auton. Robot.}, 40:429--455, 2016.

\bibitem{koptev2021real}
Mikhail Koptev, Nadia Figueroa, and Aude Billard.
\newblock \href{https://doi.org/10.1109/LRA.2021.3057024}{Real-time
  self-collision avoidance in joint space for humanoid robots}.
\newblock {\em IEEE Robot. Autom. Lett.}, 6(2):1240--1247, 2021.

\bibitem{darvish2023teleoperation}
Kourosh Darvish, Luigi Penco, Joao Ramos, and et~al.
\newblock \href{10.1109/TRO.2023.3236952}{Teleoperation of humanoid robots: A
  survey}.
\newblock {\em IEEE Trans. Robot.}, 39(3):1706--1727, 2023.

\bibitem{Kenneally2016Design}
G.~Kenneally, A.~De, and D.~E. Koditschek.
\newblock \href{https://doi.org/10.1109/LRA.2016.2528294}{Design principles for
  a family of direct-drive legged robots}.
\newblock {\em IEEE Robot. Autom. Lett.}, 1(2):900--907, 2016.

\bibitem{park2017high}
Hae~Won Park, Patrick~M Wensing, and Sangbae Kim.
\newblock \href{https://doi.org/10.1177/0278364917694244}{High-speed bounding
  with the MIT cheetah 2: Control design and experiments}.
\newblock {\em Int. J. Robot. Res.}, 36(2):167--192, 2017.

\bibitem{Kau2019Stanford}
N.~Kau, A.~Schultz, N.~Ferrante, and P.~Slade.
\newblock \href{https://doi.org/10.1109/ICRA.2019.8794436}{Stanford doggo: An
  open-source, quasi-direct-drive quadruped}.
\newblock {\em In: Proc. IEEE ICRA}, pages 6309--6315, 2019.

\bibitem{liu2020articulated}
Yujiong Liu and Pinhas Ben-Tzvi.
\newblock \href{https://doi.org/10.1115/1.4045689}{An articulated closed
  kinematic chain planar robotic leg for high-speed locomotion}.
\newblock {\em ASME J. Mech. Robot.}, 12(4):041003, 2020.

\bibitem{wu2021design}
Jianxu Wu, Long Guo, Shaoze Yan, and et~al.
\newblock \href{https://doi.org/10.1016/j.mechmachtheory.2021.104444}{Design
  and performance analysis of a novel closed-chain elastic-bionic leg with one
  actuated degree of freedom}.
\newblock {\em Mech. Mach. Theory}, 165:104444, 2021.

\bibitem{nasonov2023computational}
Kirill~V Nasonov, Dmitriy~V Ivolga, Ivan~I Borisov, and et~al.
\newblock \href{https://doi.org/10.1109/ICRA48891.2023.10161209}{Computational
  design of closed-chain linkages: Hopping robot driven by morphological
  computation}.
\newblock {\em In: Proc. IEEE ICRA}, pages 7419--7425, 2023.

\bibitem{he2020mechanism}
Jun He and Feng Gao.
\newblock \href{https://doi.org/10.1186/s10033-020-00485-9}{Mechanism,
  actuation, perception, and control of highly dynamic multilegged robots: A
  review}.
\newblock {\em Chinese. J. Mech. Eng.}, 33:1--30, 2020.

\bibitem{ha2016task}
Sehoon Ha, Stelian Coros, Alexander Alspach, and et~al.
\newblock \href{https://doi.org/10.1109/IROS.2016.7759324}{Task-based limb
  optimization for legged robots}.
\newblock {\em In: Proc. IEEE/RSJ IROS}, pages 2062--2068, 2016.

\bibitem{chadwick2020vitruvio}
Michael Chadwick, Hendrik Kolvenbach, Fabio Dubois, and et~al.
\newblock \href{https://doi.org/10.1109/LRA.2020.3013913}{Vitruvio: An
  open-source leg design optimization toolbox for walking robots}.
\newblock {\em IEEE Robot. Autom. Lett.}, 5(4):6318--6325, 2020.

\bibitem{ha2018computational}
Sehoon Ha, Stelian Coros, Alexander Alspach, and et~al.
\newblock \href{https://doi.org/10.1109/TRO.2018.2830419}{Computational design
  of robotic devices from high-level motion specifications}.
\newblock {\em IEEE Trans. Robot.}, 34(5):1240--1251, 2018.

\bibitem{zhou2022review}
Chengmin Zhou, Bingding Huang, and Pasi Fr{\"a}nti.
\newblock \href{https://doi.org/10.1007/s10845-021-01867-z}{A review of motion
  planning algorithms for intelligent robots}.
\newblock {\em J. Intell. Manuf.}, 33(2):387--424, 2022.

\bibitem{roussel2018exploratory}
Robin Roussel, Marie~Paule Cani, Jean~Claude L{\'e}on, and et~al.
\newblock \href{https://doi.org/10.1016/j.cag.2018.05.023}{Exploratory design
  of mechanical devices with motion constraints}.
\newblock {\em Comput. \& Graph.}, 74:244--256, 2018.

\bibitem{ceylan2013designing}
Duygu Ceylan, Wilmot Li, Niloy~J Mitra, and et~al.
\newblock \href{https://doi.org/10.1145/2508363.2508400}{Designing and
  fabricating mechanical automata from mocap sequences}.
\newblock {\em ACM Trans. Graph.}, 32(6):1--11, 2013.

\bibitem{coros2013computational}
Stelian Coros, Bernhard Thomaszewski, Gioacchino Noris, and et~al.
\newblock \href{https://doi.org/10.1145/2461912.2461953}{Computational design
  of mechanical characters}.
\newblock {\em ACM Trans. Graph.}, 32(4):1--12, 2013.

\bibitem{mannhart2020cami}
Dominik Mannhart, Fabio Dubois, Karen Bodie, and et~al.
\newblock \href{https://doi.org/10.1109/ICRA40945.2020.9197019}{CAMI-analysis,
  design and realization of a force-compliant variable cam system}.
\newblock {\em In: Proc. IEEE ICRA}, pages 850--856, 2020.

\bibitem{saranli2001rhex}
Uluc Saranli, Martin Buehler, and Daniel~E Koditschek.
\newblock \href{https://doi.org/10.1177/02783640122067570}{RHex: A simple and
  highly mobile hexapod robot}.
\newblock {\em Int. J. Robot. Res.}, 20(7):616--631, 2001.

\bibitem{altendorfer2001rhex}
Richard Altendorfer, Ned Moore, Haldun Komsuoglu, and et~al.
\newblock \href{https://doi.org/10.1023/A:1012426720699}{Rhex: A biologically
  inspired hexapod runner}.
\newblock {\em Auton. Robot.}, 11:207--213, 2001.

\bibitem{saranli2000design}
Uluc Saranli, Martin Buehler, and Daniel~E Koditschek.
\newblock \href{https://doi.org/10.1109/ROBOT.2000.846418}{Design, modeling and
  preliminary control of a compliant hexapod robot}.
\newblock {\em In: Proc. IEEE ICRA}, pages 2589--2596, 2000.

\bibitem{prahacs2004towards}
Chris Prahacs, Aaron Saudners, Matthew~K Smith, and et~al.
\newblock \href{https://doi.org/10.24908/pceea.v0i0.4043}{Towards legged
  amphibious mobile robotics}.
\newblock {\em In: Proc. CEEA}, 2004.

\bibitem{lin2004toward}
Pei~Chun Lin, Haldun Komsuoglu, and Daniel~E Koditschek.
\newblock \href{https://doi.org/10.1109/IROS.2004.1389746}{Toward a 6 dof body
  state estimator for a hexapod robot with dynamical gaits}.
\newblock {\em In: Proc. IEEE/RSJ IROS}, 3:2265--2270, 2004.

\bibitem{carvalho2016motion}
Joao~CM Carvalho and Tadeu~R Silvestre.
\newblock \href{https://doi.org/10.1080/15397734.2015.1051229}{Motion analysis
  of a six-legged robot using bennett's linkage as leg}.
\newblock {\em Mech. Based. Des. Stru. Mach.}, 44(1-2):86--95, 2016.

\bibitem{Gu2022OverconstrainedCoaxial}
Yuping Gu, Shihao Feng, Yuqin Guo, and et~al.
\newblock
  \href{https://doi.org/10.1016/j.mechmachtheory.2022.105018}{Overconstrained
  coaxial design of robotic legs with omni-directional locomotion}.
\newblock {\em Mech. Mach. Theory}, 176:105018, 2022.

\bibitem{gu2023computational}
Yuping Gu, Ziqian Wang, Shihao Feng, and et~al.
\newblock \href{https://doi.org/10.1093/jcde/qwad083}{Computational design
  towards energy efficient optimization in overconstrained robotic limbs}.
\newblock {\em J. Comput. Des. Eng.}, 10(5):1941--1956, 2023.

\bibitem{Sun2024OCLocomotion}
Haoran Sun, Shihao Feng, Bangchao Huang, and et~al.
\newblock \href{https://isrr2024.su.domains/}{Overconstrained locomotion}.
\newblock {\em In Proc. ISRR}, 2024.

\bibitem{song2012family}
Chaoyang Song and Yan Chen.
\newblock \href{https://doi.org/10.1098/rspa.2011.0345}{A family of mixed
  double-goldberg 6R linkages}.
\newblock {\em In: Proc. Math. Phys. Eng. Sci.}, 468(2139):871--890, 2012.

\bibitem{acharyya2009performance}
SK~Acharyya and M~Mandal.
\newblock
  \href{https://doi.org/10.1016/j.mechmachtheory.2009.03.003}{Performance of
  EAs for four-bar linkage synthesis}.
\newblock {\em Mech. Mach. Theory}, 44(9):1784--1794, 2009.

\bibitem{antonsson2001formal}
Erik~K Antonsson and Jonathan Cagan.
\newblock {\em \href{https://dl.acm.org/doi/abs/10.5555/583783}{Formal
  engineering design synthesis}}.
\newblock Cambridge University Press, 2001.

\bibitem{zhixing2002study}
Wang Zhixing, Yu~Hongying, Tang Dewei, and et~al.
\newblock \href{https://doi.org/10.1016/S0094-114X(02)00014-9}{Study on
  rigid-body guidance synthesis of planar linkage}.
\newblock {\em Mech. Mach. Theory}, 37(7):673--684, 2002.

\bibitem{nishida2019multi}
Gen Nishida, Adrien Bousseau, and Daniel~G Aliaga.
\newblock \href{https://doi.org/10.1111/cgf.13637}{Multi-pose interactive
  linkage design}.
\newblock {\em Comput. Graph. Forum}, 38(2):277--289, 2019.

\bibitem{roussel2017spirou}
Robin Roussel, Marie~Paule Cani, Jean~Claude L{\'e}on, and et~al.
\newblock \href{https://doi.org/10.1145/3083157.3083158}{SPIROU: Constrained
  exploration for mechanical motion design}.
\newblock {\em Ann. ACM Symp. Comput. Fabr.}, pages 1--11, 2017.

\bibitem{thomaszewski2014computational}
Bernhard Thomaszewski, Stelian Coros, Damien Gauge, and et~al.
\newblock \href{https://doi.org/10.1145/2601097.2601143}{Computational design
  of linkage-based characters}.
\newblock {\em ACM Trans. Graph.}, 33(4):1--9, 2014.

\bibitem{bacher2015linkedit}
Moritz B{\"a}cher, Stelian Coros, and Bernhard Thomaszewski.
\newblock \href{https://doi.org/10.1145/2766985}{Linkedit: Interactive linkage
  editing using symbolic kinematics}.
\newblock {\em ACM Trans. Graph.}, 34(4):1--8, 2015.

\bibitem{bharaj2015computational}
Gaurav Bharaj, Stelian Coros, Bernhard Thomaszewski, and et~al.
\newblock \href{https://doi.org/10.1145/2786784.2786803}{Computational design
  of walking automata}.
\newblock {\em ACM SIGGRAPH}, pages 93--100, 2015.

\bibitem{cheng2021spatial}
Yingjie Cheng, Yucheng Sun, Peng Song, and et~al.
\newblock \href{https://doi.org/10.1145/3478513.3480477}{Spatial-temporal
  motion control via composite cam-follower mechanisms}.
\newblock {\em ACM Trans. Graph.}, 40(6):1--15, 2021.

\bibitem{nollexa1975linkage}
H~Nollexa.
\newblock \href{https://doi.org/10.1016/0094-114X(75)90056-7}{Linkage coupler
  curve synthesis: A historical review — III. Spatial synthesis and
  optimization}.
\newblock {\em Mech. Mach. Theory}, 10(1):41--55, 1975.

\bibitem{cervantes2011synthesis}
J~Jes{\'u}s Cervantes-S{\'a}nchez, Luis Gracia, Eduardo Alba-Ruiz, and
  Jos{\'e}~M Rico-Mart{\'\i}nez.
\newblock \href{https://doi.org/10.1016/j.mechmachtheory.2010.10.006}{Synthesis
  of a special RPSPR spatial linkage function generator for six precision
  points}.
\newblock {\em Mech. Mach. Theory}, 46(2):83--96, 2011.

\bibitem{lin2018geometric}
Song Lin, Hanchao Wang, Jingshuai Liu, and et~al.
\newblock \href{https://doi.org/10.1115/1.4040171}{Geometric method of spatial
  linkages synthesis for function generation with three finite positions}.
\newblock {\em ASME J. Mech. Des.}, 140(8):082303, 2018.

\bibitem{bai2022exact}
Shaoping Bai, Zhongyi Li, and Jorge Angeles.
\newblock \href{https://doi.org/10.1115/1.4052336}{Exact path synthesis of RCCC
  linkages for a maximum of nine prescribed positions}.
\newblock {\em ASME J. Mech. Robot.}, 14(2):021011, 2022.

\bibitem{liu2020synthesis}
Wenrui Liu, Jianwei Sun, and Jinkui Chu.
\newblock \href{https://doi.org/10.1115/1.4044110}{Synthesis of a spatial RRSS
  mechanism for path generation using the numerical atlas method}.
\newblock {\em ASME J. Mech. Des.}, 142(1):012303, 2020.

\bibitem{cheng2022exact}
Yingjie Cheng, Peng Song, Yukun Lu, and et~al.
\newblock \href{https://doi.org/10.1145/3550454.3555431}{Exact 3D path
  generation via 3D cam-linkage mechanisms}.
\newblock {\em ACM Trans. Graph.}, 41(6):1--13, 2022.

\bibitem{sigmund2013topology}
Ole Sigmund and Kurt Maute.
\newblock \href{https://doi.org/10.1007/s00158-013-0978-6}{Topology
  optimization approaches: A comparative Review}.
\newblock {\em Stru. Multi. Opt.}, 48(6):1031--1055, 2013.

\bibitem{zhou2014boxelization}
Yahan Zhou, Shinjiro Sueda, Wojciech Matusik, and et~al.
\newblock \href{https://doi.org/10.1145/2601097.2601173}{Boxelization: Folding
  3D objects into boxes}.
\newblock {\em ACM Trans. Graph.}, 33(4):1--8, 2014.

\bibitem{yuan2018computational}
Ye~Yuan, Changxi Zheng, and Stelian Coros.
\newblock \href{https://doi.org/10.1111/cgf.13516}{Computational design of
  transformables}.
\newblock {\em Comput. Graph. Forum}, 37(8):103--113, 2018.

\bibitem{huang2016making}
Yi~Jheng Huang, Shu~Yuan Chan, Wen~Chieh Lin, and et~al.
\newblock \href{https://doi.org/10.1016/j.cag.2015.07.014}{Making and animating
  transformable 3D models}.
\newblock {\em Comput. \& Graph.}, 54:127--134, 2016.

\bibitem{anvari2018collision}
Zolfa Anvari, Parnyan Ataei, and Mehdi Tale~Masouleh.
\newblock \href{https://doi.org/10.1007/978-3-319-60867-9_41}{The
  collision-free workspace of the tripteron parallel robot based on a
  geometrical approach}.
\newblock {\em In: Proc. Int. Work. Comput. Kin.}, pages 357--364, 2018.

\bibitem{li2020invertible}
Zijia Li, Georg Nawratil, Florian Rist, and et~al.
\newblock \href{https://doi.org/10.1111/cgf.13928}{Invertible paradoxic loop
  structures for transformable design}.
\newblock {\em Comput. Graph. Forum}, 39(2):261--275, 2020.

\bibitem{zheng2016deployable}
Changxi Zheng, Timothy Sun, and Xiang Chen.
\newblock \href{https://dl.acm.org/doi/10.5555/2982818.2982843}{Deployable 3D
  linkages with collision avoidance}.
\newblock {\em In: Proc. Symp. Comput. Anim.}, 179:188, 2016.

\bibitem{kaldor2008simulating}
Jonathan~M. Kaldor, Doug~L. James, and Steve Marschner.
\newblock \href{https://doi.org/10.1145/1399504.1360664}{Simulating knitted
  cloth at the yarn level}.
\newblock {\em ACM SIGGRAPH}, pages 1--9, 2008.

\bibitem{rakita2018relaxedik}
Daniel Rakita, Bilge Mutlu, and Michael Gleicher.
\newblock \href{http://dx.doi.org/10.15607/RSS.2018.XIV.043}{RelaxedIK:
  Real-time synthesis of accurate and feasible robot arm motion}.
\newblock {\em Robot. Sci. Syst.}, 14:26--30, 2018.

\bibitem{khatib1986real}
Oussama Khatib.
\newblock \href{https://doi.org/10.1177/027836498600500106}{Real-time obstacle
  avoidance for manipulators and mobile robots}.
\newblock {\em Int. J. Robot. Res.}, 5(1):90--98, 1986.

\bibitem{nakamura1990advanced}
Yoshihiko Nakamura.
\newblock {\em \href{https://dl.acm.org/doi/abs/10.5555/533662}{Advanced
  robotics: Redundancy and optimization}}.
\newblock Addison-Wesley Longman Publishing, 1990.

\bibitem{bergou2010discrete}
Mikl{\'o}s Bergou, Basile Audoly, Etienne Vouga, and et~al.
\newblock \href{https://doi.org/10.1145/1778765.1778853}{Discrete viscous
  threads}.
\newblock {\em ACM Trans. Graph.}, 29(4):1--10, 2010.

\bibitem{wachter2006implementation}
Andreas W{\"a}chter and Lorenz~T Biegler.
\newblock \href{https://doi.org/10.1007/s10107-004-0559-y}{On the
  implementation of an interior-point filter line-search algorithm for
  large-scale nonlinear programming}.
\newblock {\em Math. Prog.}, 106:25--57, 2006.

\bibitem{andersson2019casadi}
Joel~AE Andersson, Joris Gillis, Greg Horn, and et~al.
\newblock \href{https://doi.org/10.1007/s12532-018-0139-4}{CasADi: A software
  framework for nonlinear optimization and optimal control}.
\newblock {\em Math. Prog. Comput.}, 11:1--36, 2019.

\bibitem{gogu2005chebychev}
Grigore Gogu.
\newblock
  \href{https://doi.org/10.1016/j.euromechsol.2004.12.003}{Chebychev--Gr{\"u}bler--Kutzbach's
  criterion for mobility calculation of multi-Loop mechanisms revisited via
  theory of linear transformations}.
\newblock {\em Eu. J. Mech.}, 24(3):427--441, 2005.

\bibitem{hamann2011line}
Marco Hamann.
\newblock
  \href{https://doi.org/10.1016/j.mechmachtheory.2011.02.004}{Line-symmetric
  motions with respect to reguli}.
\newblock {\em Mech. Mach. Theory}, 46(7):960--974, 2011.

\bibitem{Baker1979TheBennett}
J.~E. Baker.
\newblock \href{https://doi.org/10.1016/0094-114X(79)90011-9}{The bennett,
  goldberg and myard linkages - in perspective}.
\newblock {\em Mech. Mach. Theory}, 14(4):239--253, 1979.

\bibitem{Merlet2006Jacobian}
J.~P. Merlet.
\newblock {\href{https://doi.org/10.1115/1.2121740}{Jacobian, manipulability,
  condition number, and accuracy of parallel robots}}.
\newblock {\em ASME J. Mech. Des.}, 128(1):199--206, 2006.

\bibitem{alt1995computing}
Helmut Alt and Michael Godau.
\newblock \href{https://doi.org/10.1142/S0218195995000064}{Computing the
  fr{\'e}chet distance between two polygonal curves}.
\newblock {\em Int. J. Comput. Geom. \& Appl.}, 5(01n02):75--91, 1995.

\bibitem{rodriguez20043}
Wladimir Rodriguez, Mark Last, Abraham Kandel, and et~al.
\newblock \href{https://doi.org/10.1016/j.robot.2004.09.004}{3-Dimensional
  curve similarity using string matching}.
\newblock {\em Robot. Auton. Syst.}, 49(3-4):165--172, 2004.

\bibitem{hansen2003reducing}
Nikolaus Hansen, Sibylle~D M{\"u}ller, and Petros Koumoutsakos.
\newblock \href{https://doi.org/10.1162/106365603321828970}{Reducing the time
  complexity of the derandomized evolution strategy with covariance matrix
  adaptation (CMA-ES)}.
\newblock {\em Evolut. Comput.}, 11(1):1--18, 2003.

\bibitem{megaro2017computational}
Vittorio Megaro, Jonas Zehnder, Moritz B{\"a}cher, and et~al.
\newblock \href{https://doi.org/10.1145/3072959.3073636}{A computational design
  tool for compliant mechanisms}.
\newblock {\em ACM Trans. Graph.}, 36(4):82--1, 2017.

\bibitem{cruse2009principles}
Holk Cruse, Volker D{\"u}rr, Malte Schilling, and et~al.
\newblock \href{https://doi.org/10.1007/978-3-540-88464-4_2}{Principles of
  insect locomotion}.
\newblock {\em In: Spat. Temp. Patt. Act. Per. Rov. Rob.}, pages 43--96, 2009.

\bibitem{cruse1995movement}
Holk Cruse and Ch~Bartling.
\newblock \href{https://doi.org/10.1016/0022-1910(95)00032-P}{Movement of joint
  angles in the legs of a walking insect, carausius morosus}.
\newblock {\em J. Ins. Phys.}, 41(9):761--771, 1995.

\bibitem{cruse2006control}
Holk Cruse, Volker D{\"u}rr, Josef Schmitz, and et~al.
\newblock \href{https://doi.org/10.1007/4-431-31381-8_3}{Control of hexapod
  walking in biological systems}.
\newblock {\em Adap. Mot. Ani. Mach.}, pages 17--29, 2006.

\bibitem{garrido2014automatic}
Sergio Garrido-Jurado, Rafael Mu{\~n}oz-Salinas, Francisco~Jos{\'e}
  Madrid-Cuevas, and et~al.
\newblock \href{https://doi.org/10.1016/j.patcog.2014.01.005}{Automatic
  generation and detection of highly reliable fiducial markers under
  occlusion}.
\newblock {\em Patt. Recogn.}, 47(6):2280--2292, 2014.

\bibitem{gabrielli1950price}
Giuseppi Gabrielli.
\newblock \href{https://gwern.net/doc/technology/1950-gabrielli.pdf}{What price
  speed? Specific power required for propulsion of vehicles}.
\newblock {\em Mech. Eng.}, 72(10):775--781, 1950.

\bibitem{arena2006realization}
Paolo Arena, Luigi Fortuna, Mattia Frasca, and et~al.
\newblock \href{https://doi.org/10.1109/ISCAS.2006.1693168}{Realization of a
  CNN-driven cockroach-inspired robot}.
\newblock {\em In: Proc. IEEE Int. Symp. Circ. Syst.}, pages 4--8, 2006.

\bibitem{birkmeyer2009dash}
Paul Birkmeyer, Kevin Peterson, and Ronald~S Fearing.
\newblock \href{https://doi.org/10.1109/IROS.2009.5354561}{DASH: A dynamic 16g
  hexapedal robot}.
\newblock {\em In: Proc. IEEE/RSJ IROS}, pages 2683--2689, 2009.

\bibitem{baisch2010biologically}
Andrew~T Baisch, Pratheev~S Sreetharan, and Robert~J Wood.
\newblock
  \href{https://doi.org/10.1109/IROS.2010.5651789}{Biologically-inspired
  locomotion of a 2g hexapod robot}.
\newblock {\em In: Proc. IEEE/RSJ IROS}, pages 5360--5365, 2010.

\bibitem{xiong2015adaptive}
Xiaofeng Xiong, Florentin W{\"o}rg{\"o}tter, and Poramate Manoonpong.
\newblock \href{https://doi.org/10.1109/TCYB.2015.2479237}{Adaptive and energy
  efficient walking in a hexapod robot under neuromechanical control and
  sensorimotor learning}.
\newblock {\em IEEE Trans. Cyber.}, 46(11):2521--2534, 2015.

\bibitem{bjelonic2018weaver}
Marko Bjelonic, Navinda Kottege, Timon Homberger, and et~al.
\newblock \href{https://doi.org/10.1002/rob.21795}{Weaver: Hexapod robot for
  autonomous navigation on unstructured terrain}.
\newblock {\em J. Field Robot.}, 35(7):1063--1079, 2018.

\bibitem{luneckas2021hexapod}
Mindaugas Luneckas, Tomas Luneckas, Jonas Kriau{\v{c}}i{\=u}nas, and et~al.
\newblock \href{https://doi.org/10.3390/app11031339}{Hexapod robot gait
  switching for energy consumption and cost of transport management using
  heuristic algorithms}.
\newblock {\em Appl. Sci.}, 11(3):1339, 2021.

\bibitem{xu2020computational}
Hao Xu, Tianwen Fu, Peng Song, and et~al.
\newblock \href{https://doi.org/10.1111/cgf.13939}{Computational design and
  optimization of non-circular gears}.
\newblock {\em Comput. Graph. Forum}, 39(2):399--409, 2020.

\bibitem{shin2023topology}
Seungyeon Shin, Dongju Shin, and Namwoo Kang.
\newblock \href{https://doi.org/10.1093/jcde/qwad072}{Topology optimization via
  machine learning and deep learning: A review}.
\newblock {\em J. Comput. Des. Eng.}, 10(4):1736--1766, 2023.

\bibitem{abas2016flapping}
Mohd Firdaus~Bin Abas, Azmin Shakrine Bin~Mohd Rafie, Hamid~Bin Yusoff, and
  et~al.
\newblock \href{https://doi.org/10.1016/j.cja.2016.08.003}{Flapping wing
  micro-aerial-vehicle: Kinematics, membranes, and flapping mechanisms of
  ornithopter and insect flight}.
\newblock {\em Chinese. J. Aeron.}, 29(5):1159--1177, 2016.

\bibitem{khan2018design}
Masood~M Khan and Cheng Chen.
\newblock \href{https://doi.org/10.1109/HUMANOIDS.2018.8625069}{Design of a
  single cam single actuator multiloop eyeball mechanism}.
\newblock {\em In: Proc. IEEE/RAS Int. Conf. Human. Robot.}, pages 1143--1149,
  2018.

\end{thebibliography}
\end{document}